\documentclass{article}
\usepackage[utf8]{inputenc}
\usepackage[a4paper, total={6in, 8in}]{geometry}
\usepackage[round]{natbib}
\usepackage{amsmath}
\usepackage{amssymb}
\usepackage{pgfplots}
\pgfplotsset{compat=1.15}
\usetikzlibrary{arrows}
\usetikzlibrary {arrows.meta}
\usetikzlibrary{math}
\usetikzlibrary{positioning}
\usetikzlibrary{decorations.pathreplacing}

\tikzstyle{state}=[circle, draw=black, fill=black]
\tikzstyle{action}=[circle, draw=black, fill=white]
\tikzstyle{reward}=[circle,  fill=black!50]
\tikzstyle{rollout}=[-]
\tikzstyle{ml_rollout}=[Turned Square-{Turned Square[open]}]
\tikzstyle{transition_func}=[->]
\tikzstyle{reward_func}=[-{Triangle[open, length=3mm, width=2mm]}]
\tikzstyle{statemap}=[Turned Square, draw=black, fill=black]
\tikzstyle{actionmap}=[Turned Square, draw=black, fill=white]

\newcommand{\rollout}[2]{\tikzmath{\x = #1; \y =#2;};
\node[state](s) at (\x,\y) {};
\node[action](a) at (\x+1,\y) {};
\node[reward](r) at (\x+1,\y+1) {};
\draw[rollout] (s) -- (a);
\draw[reward_func] (a) -- (r);}

\newcommand{\rollouttrans}[2]{\tikzmath{\x = #1; \y =#2;};
\node[state](s) at (\x,\y) {};
\node[action](a) at (\x+1,\y) {};
\node[reward](r) at (\x+1,\y+1) {};
\draw[rollout] (s) -- (a);
\draw[reward_func] (a) -- (r);
\draw[transition_func] (a) -- (\x+1.8,\y);}

\newcommand{\rolloutdots}[2]{\tikzmath{\x = #1; \y =#2;};
\node[state](s) at (\x,\y) {};
\node[action](a) at (\x+1,\y) {};
\node[reward](r) at (\x+1,\y+1) {};
\draw[rollout] (s) -- (a);
\draw[reward_func] (a) -- (r);
\node at (\x+1.5, \y) {$\boldsymbol{\cdots}$};}

\newcommand{\rdots}[2]{\tikzmath{\x = #1; \y =#2;};
\node at (\x+1.5, \y) {$\boldsymbol{\cdots}$};}

\newcommand{\mlrollout}[2]{\tikzmath{\x = #1; \y =#2;};
\node[state](s) at (\x,\y) {};
\node[action](a) at (\x+1,\y) {};
\node[reward](r) at (\x+1,\y+1) {};
\draw[ml_rollout] (s) -- (a);
\draw[reward_func] (a) -- (r);}

\newcommand{\mlrollouttrans}[2]{\tikzmath{\x = #1; \y =#2;};
\node[state](s) at (\x,\y) {};
\node[action](a) at (\x+1,\y) {};
\node[reward](r) at (\x+1,\y+1) {};
\draw[ml_rollout] (s) -- (a);
\draw[reward_func] (a) -- (r);
\draw[transition_func] (a) -- (\x+1.8,\y);}

\newcommand{\mlrolloutdots}[2]{\tikzmath{\x = #1; \y =#2;};
\node[state](s) at (\x,\y) {};
\node[action](a) at (\x+1,\y) {};
\node[reward](r) at (\x+1,\y+1) {};
\draw[ml_rollout] (s) -- (a);
\draw[reward_func] (a) -- (r);
\node at (\x+1.5, \y) {$\boldsymbol{\cdots}$};}
\usepackage{algorithm} 
\usepackage{algpseudocode}
\usepackage{subfig}
\usepackage{lineno}
\usepackage{lipsum} 
\usepackage{url}
\usepackage[colorlinks=true,citecolor=blue,filecolor=magenta,urlcolor=red]{hyperref}
\usepackage{authblk}

\algdef{SE}[SUBALG]{Indent}{EndIndent}{}{\algorithmicend\ }%
\algtext*{Indent}
\algtext*{EndIndent}

\title{\textbf{A Multilevel Reinforcement Learning Framework for PDE-based Control} }
\author{Atish Dixit, Ahmed H. Elsheikh}
\affil{Herriot-Watt University,\\ Edinburgh, UK.}
\date{\today}

\begin{document} \sloppy

\maketitle

\begin{abstract}
Reinforcement learning (RL) is a promising method to solve control problems~\citep{Atish2022a}. However, model-free RL algorithms are sample inefficient and require thousands if not millions of samples to learn optimal control policies. A major source of computational cost in RL corresponds to the transition function, which is dictated by the model dynamics. This is especially problematic when model dynamics is represented with coupled PDEs. In such cases, the transition function often involves solving a large-scale discretization of the said PDEs. We propose a multilevel RL framework in order to ease this cost by exploiting sublevel models that correspond to coarser scale discretization (i.e. multilevel models). This is done by formulating an approximate multilevel Monte Carlo estimate (inspired by \cite{giles2015multilevel}) of the objective function of the policy and / or value network instead of Monte Carlo estimates, as done in the classical framework. As a demonstration of this framework, we present a multilevel version of the proximal policy optimization (PPO) algorithm. Here, the level refers to the grid fidelity of the chosen simulation-based environment. We provide two examples of simulation-based environments that employ stochastic PDEs that are solved using finite-volume discretization. For the case studies presented, we observed substantial computational savings using multilevel PPO compared to its classical counterpart.
\end{abstract}

\section{Introduction} \label{sec:level1}
Optimal control problem involves finding controls for a dynamical system (often represented by a set of partial differential equations (PDEs)) such that a certain objective function is optimized over a predefined simulation time. 
In recent years, we have seen a surge in research activities where reinforcement learning (RL) has been demonstrated as an effective method to solve optimal control problems in fields such as energy \citep{anderlini2016control}, fluid dynamics \citep{rabault2019artificial}, and subsurface flow control \citep{Atish2022a}. 
The reinforcement learning process for optimal control policy often involves a large number of exploration and exploitation attempts of control trajectories. 
In the context of PDE-based control problems, this corresponds to a large number of simulations of the underlying model dynamics.
For large-scale PDE-based problems (i.e., with high-fidelity PDE discretization), this makes RL a computationally expensive process.

Since the introduction of the multilevel Monte Carlo (MLMC) estimate as a computationally cheaper counterpart to classical Monte Carlo estimates, numerous research studies have been conducted in the application of MLMC estimates in uncertainty quantification for stochastic PDEs \citep{cliffe2011multilevel,anderson2012multilevel,giles2018multilevel}. 
Furthermore, we also see a rise of MLMC estimate applications in certain deep learning research studies.
For example, \citet{shi2021multilevel} present a framework for MLMC-based unbiased gradient estimation in deep latent variable models.
\citet{chada2022multilevel} illustrate how the MLMC method could be applied to Bayesian inference using deep neural networks to compute expectations associated with the posterior distribution where the level corresponds to the sets of neural network parameters under consideration.
In this paper, we introduce a novel multilevel framework for reinforcement learning where the learned agent interacts with environments corresponding to simulations of PDEs and the level corresponds to the grid fidelity of the PDE discretization.

We start by presenting the anatomy for classical RL algorithms, which involves estimating the Monte Carlo estimate of the objective function for policy and/or value network.
Furthermore, we formulate the approximate MLMC estimation methodology used in the proposed multilevel framework.
We then briefly present the mathematical framework that enables synchronized rollouts of task trajectories at different levels of the environment. 
The data generated through these synchronized rollouts are used to compute the approximate MLMC estimate of the objective function. 
Using the proposed multilevel framework, we formulate a multilevel variant of the state-of-the-art algorithm titled proximal policy optimization (PPO).

In the experiments presented, we compare the reinforcement learning process for classical and proposed multilevel PPO algorithms.
The results are demonstrated for two environments for which the model dynamics is represented by stochastic partial differential equations.
Furthermore, we also demonstrate the results of standard MLMC analysis to compare the MLMC and MC estimates for the PPO objective function.
These environments were inspired by our research work in \citet{Atish2022a}.
In this study, the levels of the environment correspond to the discretization fidelity of the grid of the underlying PDEs. 

The following is the outline for the rest of the paper: Section \ref{sect: background} provides the anatomy of the classical RL framework and formally defines the approximate MLMC estimation method.
Section \ref{sect: ml_framework} introduces the multilevel framework for RL algorithms and further presents the multilevel PPO algorithm along with its analysis methodology. 
Numerical experiments to demonstrate the proposed multilevel PPO algorithm are detailed in Section \ref{sect: experiments}, and the results of these experiments are delineated in Section \ref{sect: results}. 
Finally, Section \ref{sect: conclusion} concludes with a summary of the research study and an outlook on future research directions.

\section{Background} \label{sect: background}
Conventionally, the RL framework consists of the environment $\mathcal{E}$, which is governed by a Markov decision process described by the tuple $\left \langle \mathcal{S},\mathcal{A},\mathcal{P},\mathcal{R}, \mu \right \rangle$.
Here, $\mathcal{S} \subset \mathbb{R}^{n_s}$ is the state-space, $\mathcal{A} \subset \mathbb{R}^{n_a}$ is the action-space, 
$\mathcal{P}(s'|s,a)$ is a Markov transition probability function between the current state $s$ and the next state $s'$ under action $a$ and $\mathcal{R}(s,a,s')$ is the reward function.
The function $\mu(s)$ returns a state from the initial state distribution if $s$ is the terminal state of the episode (e.g., simulation terminal time); otherwise, it returns the same state $s$. 
The goal of reinforcement learning is to find the policy $\pi(a|s)$ to take an optimal action $a$ when the state $s$ is observed.
In deep reinforcement learning, the policy is denoted $\pi_\theta(a|s)$ and is represented by a neural network with parameters $\theta$ either directly (for policy-based algorithms) or indirectly (for value-based algorithms). Learning is initiated with a random policy and then updated by exploring state-action spaces and exploiting the observed rewards in subsequent sampling steps. Each such update is referred to as a policy iteration.
 \begin{algorithm*}
	\caption{Anatomy of deep reinforcement learning algorithms} 
	\begin{algorithmic}[1]
	\For{$\textup{policy iteration}=1,2,\ldots$}
	    \State \textbf{step 1:} Generate sequences $\{s_t,a_t,r_t\}_{t=1}^{t=T}$ using current policy $\pi_{\theta}(a|s)$
	    \Indent
	        \For {$t=1,2,\ldots, T$}
	            \State generate samples $s_t, a_t$ and $r_t$, where $s,a,r \sim p_\theta $
	            \State compute $\Theta_t$
	        \EndFor
	    \EndIndent
	    \State \textbf{step 2:} Compute Monte Carlo estimate of objective function $\mathbb{E}_{s,a,r \sim p_{\theta}} [ J (s,a,r; \theta, \Theta) ]$:
	    \Indent
	        \State $\widehat{\mathbb{E}}^T_{s,a,r\sim p_{\theta}} \left [ J(s,a,r;\theta, \Theta) \right ] $
	    \EndIndent
	    \State \textbf{step 3:} Update $\theta$ using the gradient of the estimated objective function
	    
	\EndFor
	\end{algorithmic} 
	\label{alg: classical_anatomy}
\end{algorithm*}

The algorithm \ref{alg: classical_anatomy} outlines a general anatomy of deep reinforcement learning algorithms. 
Each policy iteration consists of three steps. 
First, the sequence $\{(s_1,a_1,r_1),\ldots,(s_T, a_T, r_T)\}$ is generated by rolling out the current policy $\pi_{\theta}(a|s)$. In this stage, the RL algorithm utilizes the current policy to interact with the simulated environment by providing actions (aka. controls) and recording the observed rewards. A shorthand notation $s,a,r \sim p_{\theta}$, is used for the definition of random variables $s,a$ and $r$. Equation \ref{eq: mc_rollout_notation} provides a detailed expansion of this shorthand notation.
\begin{equation}
s,a,r \sim p_{\theta}
\left\{\begin{matrix}
s \sim \mu(s)\\ 
a \sim \pi_{\theta}(a|s) \\ 
s' \sim \mathcal{P}(s,a) \\ 
r = \mathcal{R}(s, a, s')
\end{matrix}\right.
\label{eq: mc_rollout_notation}
\end{equation}

The objective function used to calculate the gradient of the network parameters $\theta$, is of the form $\mathbb{E}_{s,a,r \sim p_{\theta}} [ J (s,a,r; \theta, \Theta) ]$, where $\Theta$ is a set of other parameters that can vary from one algorithm to another.
Appendix \ref{app: a} delineates this objective function for various algorithms.
The second step consists of computing a Monte Carlo estimate of $\mathbb{E}_{s,a,r \sim p_{\theta}} [ J (s,a,r; \theta, \Theta) ]$ which is calculated using the samples generated in the first step. 
The notation $\widehat{\mathbb{E}}^T_{x \sim \mathcal{X}} [f(x)]$ in algorithm \ref{alg: classical_anatomy} corresponds to the Monte Carlo estimate of $\mathbb{E}_{x \sim \mathcal{X}} [f(x)]$ which is calculated as $T^{-1}\sum_{t=0}^T f(x_t)$, where $x_1,\ldots,x_T$ are random samples of the random variable $x \sim \mathcal{X}$. 
To maintain brevity in the description of Monte Carlo estimate, we use the same notation in the rest of the paper.
Finally, in the third step, the policy is updated by updating the network parameters $\theta$ using the gradient of the estimated objective function.

\subsection{Approximate Multilevel Monte Carlo estimation}

Monte Carlo estimate of $\mathbb{E}[f(x^L)]$ for the random variable $x^L \sim \mathcal{X}^L$ is defined as
\begin{equation*}
    \mathbb{E}_{x^L \sim \mathcal{X}^L}[f(x^L)] \approx  \widehat{\mathbb{E}}^T_{x^L \sim \mathcal{X}^L}[f(x^L)],
\end{equation*}
where $T$ denotes the number of samples used in the estimation.
Suppose that we have functions $\varphi_L^l$ that approximate the random variable $x^L$ from level $L$ to $l$, $\forall l \in \{1,2,\ldots,L\}$ (note that $\varphi_L^L$ is simply an identity function).
Functions $\varphi_L^l$ are defined so that each decrease in level $l$ corresponds to a proportional decrease in the accuracy and cost of computing the function $f(\varphi_L^l(x^L))$.
In PDE-based uncertainty quantification problems, the function $f$ represents the quantity of interest, which implicitly contains the solution for the said PDE. 
The level refers to the grid discretization used during the PDE solving; that is, the grid discretization goes from coarsest to finest from level 1 to $L$.
For such a multilevel representation of functions, MLMC estimate of $\mathbb{E}[f(x^L)]$ is defined as
\begin{equation}
    \mathbb{E}_{x^L \sim \mathcal{X}^L}[f(x^L)] \approx  \sum_{l=1}^L \widehat{\mathbb{E}}^{T^l}_{x^L \sim \mathcal{X}^L}[f( \varphi_L^l(x^L)) - f( \varphi_L^{l-1}(x^L))],
    \label{eq: mlmc_actual}
\end{equation}
where $T^l$ represents the number of samples at each level $l$, and the value of the function at the zeroth level is predefined at zero (that is, $f(\varphi_L^0())\doteq0$). 
The MLMC estimate is introduced by \cite{giles2015multilevel} as a computationally cheaper alternative to the classical Monte Carlo estimate.
Readers are referred to the Appendix \ref{app: funda_mlmc} where we briefly explain the principle behind the computational savings in the MLMC estimation.

As described in Equation \ref{eq: mlmc_actual}, the MLMC estimate is the telescopic sum of Monte Carlo estimates of the difference term $f( \varphi_L^l(x^L)) - f( \varphi_L^{l-1}(x^L))$ $\forall l \in \{1,2,\ldots,L\}$, for the random variable $x^L \sim \mathcal{X}^L$.
We reformulate this MLMC estimate so that we can use approximate samples at each level instead of samples from the finest level $L$. 
This is done with the following two approximations:
First, we treat $\varphi_L^l(x^L)$ as a random variable $x^l \sim \mathcal{X}^l$, $\forall l \in \{1,2,\ldots,L\}$.
Second, we replace the second difference term from $\varphi_L^{l-1}(x^L)$ to $\varphi_l^{l-1}(x^l)$.
In other words, the difference term can now be computed using an approximate random variable $x^l$ as opposed to the random variable on the finest level $x^L$.
Furthermore, this term $\varphi_l^{l-1}(x^l)$, is denoted with $\tilde{x}^{l-1}$, which represents the synchronized value of $x^l$ at the level $l-1$.
We denote this synchronization process by the shorthand notation $\tilde{x}^{l-1} = \mathcal{X}^{l\Rightarrow{l-1}}$ as a subscript.
Taking these approximations into account, we formulate the approximate MLMC estimate as follows.
\begin{equation}
     \sum_{l=1}^L \widehat{\mathbb{E}}^{T^l}_{\substack{x^l \sim \mathcal{X}^l \\ \tilde{x}^{l-1} = \mathcal{X}^{l \Rightarrow l-1} }}[f(x^l) - f(\tilde{x}^{l-1} )].
     \label{eq: mlmc_est_2}
\end{equation}
Note that with this formulation, we can employ the random variable $x^l$ at each level $l$. This idea of using approximate samples at each level is at the heart of the proposed multilevel RL framework. In the rest of the paper, we use the Equation \ref{eq: mlmc_est_2} notation to formulate the approximate estimate of MLMC. 

\section{Multilevel RL framework} \label{sect: ml_framework}

We introduce a multilevel RL framework formulated as a tuple, $\left \langle \boldsymbol{\mathcal{E}}, \psi_l^{l'}, \phi_l^{l'} \right \rangle$ where $\boldsymbol{\mathcal{E}}$ represents a set of multiple environments $\{ \mathcal{E}^1, \mathcal{E}^2, \ldots, \mathcal{E}^L \}$. 
An environment $\mathcal{E}^L$ corresponds to the target task described by the tuple $\left \langle \mathcal{S}^L,\mathcal{A}^L,\mathcal{P}^L,\mathcal{R}^L, \mu^L \right \rangle$. Its corresponding sublevel tasks are represented as environments $\mathcal{E}^1, \mathcal{E}^2, \ldots, \mathcal{E}^{L}$ such that the computational cost of $\mathcal{P}^l$ and the accuracy of $\mathcal{R}^l$ is lower than $\mathcal{P}^{l+1}$ and $\mathcal{R}^{l+1}$ for all values of $l \in \{1,\ldots,L-1\}$. 
Furthermore, $\psi_l^{l'}(s^l)$ is a mapping function from state on level $l$ (denoted as $s^l$) to state on level $l'$ (denoted as $s^{l'}$) and similarly $\phi_l^{l'}(a^l)$ is a mapping function from action $a^l$ to $a^{l'}$.
The algorithm \ref{alg: multilevel_anatomy} outlines the anatomy of deep reinforcement learning algorithms with the proposed multilevel framework. 
 \begin{algorithm*}
	\caption{Anatomy for multilevel deep reinforcement learning algorithms} 
	\begin{algorithmic}[1]
	\For{$\textup{policy iteration}=1,2,\ldots$}
	    \State \textbf{step 1:} Generate sequences $\{\{(s^l_t,a^l_t,r^l_t), (\tilde{s}^{l-1}_t,\tilde{a}^{l-1}_t,\tilde{r}^{l-1}_t)\}_{t=1}^{t=T}\}_{l=1}^{l=L}$ with policy $\pi_{\theta}(a^L|s^L)$
	    \Indent
	        \For { level $l=1,2,\ldots, L$}
	           \State $s^l_t = \psi_{l-1}^l(\tilde{s}_{T_{l-1}}^{l-1})$ \algorithmiccomment{\textbf{if} $l>1$}
	            \For{$t=1,2,\ldots,T_l$}
	                \State generate samples $s^l_t,a^l_t,r^l_t$ where $s^l,a^l,r^l \sim p^l_\theta $ 
	                \State compute $\Theta^l_t$
	                \State generate synchronised samples $\tilde{s}^{l-1}_t,\tilde{a}^{l-1}_t,\tilde{r}^{l-1}_t$  \algorithmiccomment{\textbf{if} $l>1$}
	                \State where $\tilde{s}^{l-1},\tilde{a}^{l-1},\tilde{r}^{l-1} = p^{l\Rightarrow l-1}_\theta $
	                \State compute $\tilde{\Theta}^{l-1}_t$ \algorithmiccomment{\textbf{if} $l>1$}   
	            \EndFor
	        \EndFor
	    \EndIndent
	    \State \textbf{step 2:} Compute approximate multilevel Monte Carlo estimate of objective $\mathbb{E}_{s,a,r \sim p_{\theta}} [ J (s,a,r; \theta, \Theta) ]$:
	    \Indent
	        \State $ \sum_{l=1}^{l=L} \widehat{\mathbb{E}}^{T_l}_{\substack{s^l,a^l,r^l \sim p^l_\theta \\ \tilde{s}^{l-1},\tilde{a}^{l-1},\tilde{r}^{l-1} = p^{l\Rightarrow l-1}_\theta}} \left [ J(s^l,a^l,r^l;\theta, \Theta^l) - J(\tilde{s}^{l-1},\tilde{a}^{l-1},\tilde{r}^{l-1};\theta, \tilde{\Theta}^{l-1}) \right ] $,
	        \State where $J(\tilde{s}^0,\tilde{a}^0,\tilde{r}^0;\theta, \tilde{\Theta}^0_t) \doteq 0$.
	    \EndIndent
	    \State \textbf{step 3:} Update $\theta$ using the gradient of estimated objective function
	    
	\EndFor
	\end{algorithmic} 
	\label{alg: multilevel_anatomy}
\end{algorithm*}

\begin{figure}
    \centering
    \begin{tabular}{c c}
        \subfloat[symbol representation] { {\resizebox{0.4\columnwidth}{!} {\begin{tikzpicture}[line cap=round,line join=round,>=triangle 45,x=1cm,y=1cm, scale=1.0]
\clip(-1,-1.5) rectangle (6,3);


\node[state] at (0,2.5) {} (1,2.5) node[right] {state, $s$};
\node[action] at (0,2.0) {} (1,2.0)  node[right] {action, $a$};
\node[reward] at (0,1.5) {} (1,1.5) node[right] {reward, $r$};
\draw[rollout] (0,1.0) -- (0.5,1.0); 
\node[right] at (1,1.0) {policy rollout, $\pi_{\theta}(a|s)$};
\draw[transition_func] (0,0.5) -- (0.5,0.5); 
\node[right] at (1,0.5) {transition function, $\mathcal{P}(s'|s,a)$};
\draw[reward_func] (0,0) -- (0.5,0); 
\node[right] at (1,0) {reward function, $\mathcal{R}(s,a,s')$};
\draw[ml_rollout] (0,-0.5) -- (0.5,-0.5); 
\node[right] at (1,-0.5) {policy rollout $\pi_{\theta}(a^L|s^L)$, with};
\node[right] at (1,-1) {$s$ and $a$ mappings: $\psi^L_l$, $\phi^l_L$ };

\end{tikzpicture}}}
        }
        &
        \subfloat[rollouts in classical framework] {\resizebox{0.5\columnwidth}{!} {\begin{tikzpicture}[line cap=round,line join=round,>=triangle 45,x=1cm,y=1cm, scale=1.0]
\clip(-1,-1) rectangle (12,3);


\rollouttrans{0}{1}
\node at (0.6, -0.3) {step 1};
\rollouttrans{2}{1}
\node at (2.6, -0.3) {step 2};
\rolloutdots{4}{1}
\node at (4.6, -0.3) {step 3};
\node at (5.5, -0.3) {$\ldots$};
\rollouttrans{6}{1}
\node at (7, -0.3) {step $T-2$};
\rollouttrans{8}{1}
\node at (9, -0.3) {step $T-1$};
\rollout{10}{1}
\node at (11, -0.3) {step $T$};

\end{tikzpicture}}}
    \end{tabular}
    \subfloat[multilevel synchronised rollouts in proposed framework] {\resizebox{0.95\columnwidth}{!} {\begin{tikzpicture}[line cap=round,line join=round,>=triangle 45,x=1cm,y=1cm, scale=1.0]
\clip(-3,3) rectangle (26,20);


\newcommand\lvlone{17}
\newcommand\lvltwo{15}
\newcommand\lvlthree{13}
\newcommand\lvldots{11}
\newcommand\lvlsecondlast{9}
\newcommand\lvllast{7}
\newcommand\shiftmap{3}
\newcommand\syncmap{1}

\node at (-1.5, \lvlone) {\Large{level 1}};
\mlrollouttrans{0}{\lvlone}
\mlrollouttrans{2}{\lvlone}
\mlrolloutdots{4}{\lvlone}
\mlrollout{6}{\lvlone}
\mlrollout{8}{\lvlone}
\mlrolloutdots{10}{\lvlone}
\draw [decorate, ultra thick,
    decoration = {brace, amplitude=10pt}] (6,\lvlone-1) --  (0,\lvlone-1) node [black,midway,yshift=-20pt] {\Large{
$T_1$ steps}};

\node at (-1.5, \lvltwo) {\Large{level 2}};
\mlrollouttrans{6}{\lvltwo}
\mlrollouttrans{8}{\lvltwo}
\mlrolloutdots{10}{\lvltwo}
\mlrollout{12}{\lvltwo}
\mlrollout{14}{\lvltwo}
\mlrolloutdots{16}{\lvltwo}
\draw [decorate, ultra thick,
    decoration = {brace, amplitude=10pt}] (12,\lvltwo-1) --  (6,\lvltwo-1) node [black,midway,yshift=-20pt] {\Large{
    $T_2$ steps,}} node [black,midway,yshift=-40pt] {\Large{with $\tilde{s}^1=\psi_2^1(s^2)$}} node [black,midway,yshift=-60pt] {\Large{synchronisation}};

\node at (-1.5, \lvlthree) {\Large{level 3}};
\mlrollouttrans{12}{\lvlthree}
\mlrollouttrans{14}{\lvlthree}
\mlrolloutdots{16}{\lvlthree}
\draw [decorate, ultra thick,
    decoration = {brace, amplitude=10pt}] (17,\lvlthree-1) --  (12,\lvlthree-1) node [black,midway,yshift=-20pt] {\Large{
    $T_3$ steps,}} node [black,midway,yshift=-40pt] {\Large{with $\tilde{s}^2=\psi_3^2(s^3)$}} node [black,midway,yshift=-60pt] {\Large{synchronisation}};

\rdots{-2.5}{\lvldots-0.5}
\rdots{-2.5}{\lvldots}
\rdots{-2.5}{\lvldots+0.5}
\rdots{16}{\lvldots-0.5}
\rdots{16}{\lvldots}
\rdots{16}{\lvldots+0.5}

\node at (-1.5, \lvlsecondlast) {\Large{level $L-1$}};
\rdots{16}{\lvlsecondlast}
\mlrollout{18}{\lvlsecondlast}
\mlrollout{20}{\lvlsecondlast}
\mlrolloutdots{22}{\lvlsecondlast}
\mlrollout{24}{\lvlsecondlast}

\node at (-1.5, \lvllast) {\Large{level $L$}};
\rdots{16}{\lvllast}
\mlrollouttrans{18}{\lvllast}
\mlrollouttrans{20}{\lvllast}
\mlrolloutdots{22}{\lvllast}
\mlrollout{24}{\lvllast}
\draw [decorate, ultra thick,
    decoration = {brace, amplitude=10pt}] (25,\lvllast-1) --  (18,\lvllast-1) node [black,midway,yshift=-20pt] {\Large{
    $T_L$ steps,}} node [black,midway,yshift=-40pt] {\Large{with $\tilde{s}^{L-1}=\psi_L^{L-1}(s^L)$}} node [black,midway,yshift=-60pt] {\Large{synchronisation}};

\end{tikzpicture}}}
    \caption{schematics of rollouts for a policy iteration}
    \label{fig: rollouts_schema}
\end{figure}
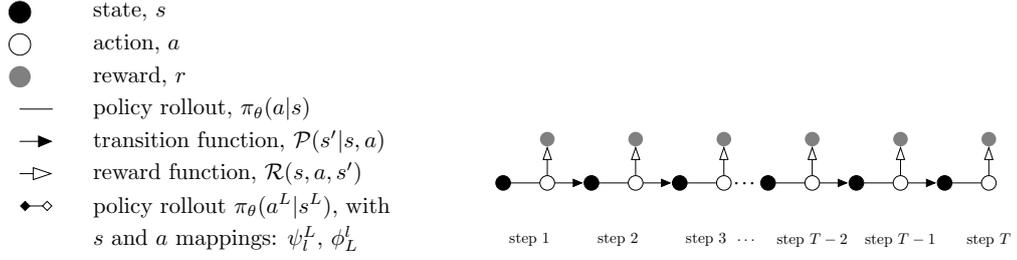
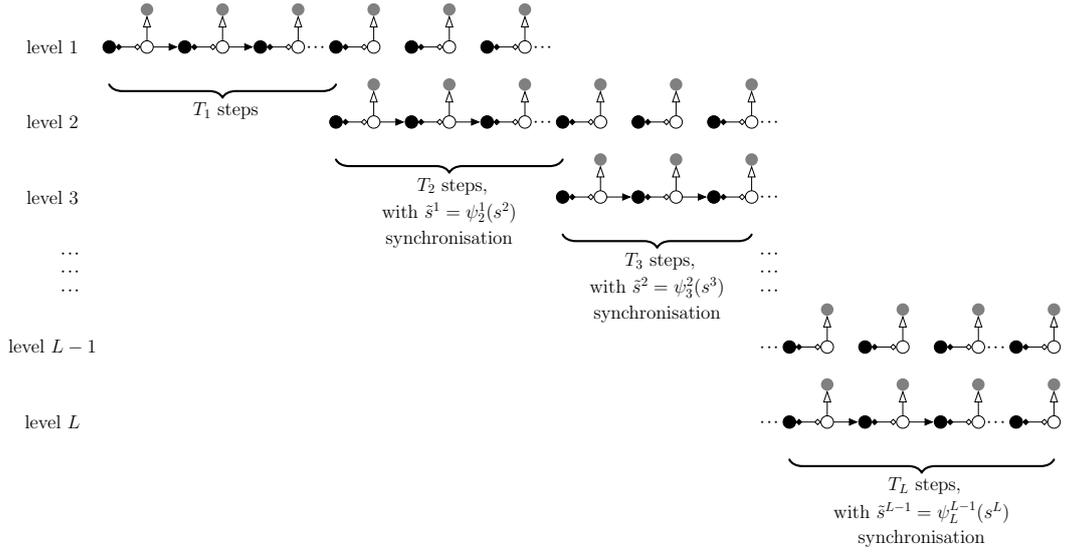

The first step consists of generating the sequence $\{(s^l_1, a^l_1, r^l_1), \ldots, (s^l_{T_l}, a^l_{T_l}, r^l_{T_l}) \}$ on level $l$ and its corresponding synchronized sequence $\{(\tilde{s}^{l-1}_1, \tilde{a}^{l-1}_1, \tilde{r}^{l-1}_1), \ldots, (\tilde{s}^{l-1}_{T_l}, \tilde{a}^{l-1}_{T_l}, \tilde{r}^{l-1}_{T_l}) \}$  on level $l-1$.
The shorthand notation $s^l,a^l,r^l \sim p^l_\theta $, for generating rollouts at level $l$ and $\tilde{s}^{l-1},\tilde{a}^{l-1},\tilde{r}^{l-1} = p^{l\Rightarrow l-1}_\theta $ for generating its synchronized rollouts at level $l-1$ are expanded in Equation \ref{eq: ml_rollout_notation}.
\begin{equation}
\begin{matrix}
s^l,a^l,r^l \sim p^l_{\theta}
\left\{\begin{matrix}
s^l \sim \mu(s^l)\\ 
s^L = \psi^L_l(s^l) \\
a^L \sim \pi_{\theta}(a^L|s^L) \\ 
a^l = \phi^l_L(a^L) \\
s'^l \sim \mathcal{P}^l(s^l,a^l) \\ 
r^l = \mathcal{R}^l(s^l, a^l, s'^l)
\end{matrix}\right. & 
\tilde{s}^{l-1},\tilde{a}^{l-1},\tilde{r}^{l-1} = p^{l \Rightarrow l-1}_{\theta}
\left\{\begin{matrix}
\tilde{s}^{l-1} = \psi^{l-1}_l(s^l)\\ 
\tilde{s}^L = \psi^L_{l-1}(\tilde{s}^{l-1}) \\
\tilde{a}^L \sim \pi_{\theta}(\tilde{a}^L|\tilde{s}^L) \\ 
\tilde{a}^{l-1} = \phi^{l-1}_L(\tilde{a}^L) \\
\tilde{s}'^{l-1} \sim \mathcal{P}^{l-1}(\tilde{s}^{l-1},\tilde{a}^{l-1}) \\ 
\tilde{r}^{l-1} = \mathcal{R}^{l-1}(\tilde{s}^{l-1}, \tilde{a}^{l-1}, \tilde{s}'^{l-1})
\end{matrix}\right.
\end{matrix}
\label{eq: ml_rollout_notation}
\end{equation}
Note that since the target task corresponds to the level $L$, the policy is now represented as $\pi_{\theta}(a^L|s^L)$.
Consequently, during policy rollouts on a certain level $l$, the state $s^l$ passes through the mapping $\psi_l^L$ and the action obtained $a^L$ passes through the mapping $\phi_L^l$. 
Synchronization from level $l$ to $l-1$ is obtained by mapping the states: $\tilde{s}^{l-1} = \psi^{l-1}_l(s^l)$.
Figure \ref{fig: rollouts_schema} illustrates the implementations of a policy iteration in the classical and multilevel frameworks.
Note that the level $l-1$ changes to $l$ at the end of steps $T_{l-1}$ (for $l=2,\ldots,L$) and to continue the rollouts at the level $l$, the state is mapped as $ \psi_{l-1}^l(\tilde{s}_{T_{l-1}}^{l-1})$. 
The generated samples are further used to compute the approximate multilevel Monte Carlo estimate of $\mathbb{E}_{s,a,r \sim p_{\theta}} [ J (s,a,r; \theta, \Theta) ]$ which is described as
\begin{align*}
         \sum_{l=1}^{l=L} \widehat{\mathbb{E}}^{T_l}_{\substack{s^l,a^l,r^l \sim p^l_\theta \\ \tilde{s}^{l-1},\tilde{a}^{l-1},\tilde{r}^{l-1} = p^{l\Rightarrow l-1}_\theta}} \left [ J(s^l,a^l,r^l;\theta, \Theta^l) - J(\tilde{s}^{l-1},\tilde{a}^{l-1},\tilde{r}^{l-1};\theta, \tilde{\Theta}^{l-1}) \right ],
\end{align*}
where $J(\tilde{s}^0,\tilde{a}^0,\tilde{r}^0;\theta, \tilde{\Theta}^0) \doteq 0$. Since $T_1>T_2>\ldots>T_L$, most of the computational costs of the rollouts lean towards sublevel environments. 
As a result, the approximate multilevel Monte Carlo estimate requires an overall lower computational cost than the Monte Carlo estimate in the classical framework. 
Finally, the network parameters $\theta$ are updated using the gradient of the estimated objective function at the end of the policy iteration.

\subsection{Multilevel PPO algorithm} \label{sect: mlppo}
We present the proposed multilevel framework for the state-of-the-art model-free algorithm, proximal policy optimization (PPO) \citep{schulman2017proximal}. 
In the context of the multilevel framework, the objective function is defined as
 \begin{align}
    \begin{split}
        J(s,a,r;\theta, \Theta^{ppo}) =& \min \left (p(\theta) A(s,a), \textup{clip} (\textbf{r}(\theta), 1-\epsilon, 1+\epsilon)A(s,a) \right )\\
     &- c_v \left (  r + \gamma \max_{s'}V_{\theta}(s') - V_{\theta}(s) \right )^2\\ 
     &+ c_e S[\pi_\theta](s).
    \end{split}
    \label{eq: ppo_obj}
 \end{align}
The first term of this objective function is called the surrogate policy term, where $p(\theta) = \pi_{\theta}(a|s)/\pi_{\theta_{old}}(a|s)$ and $\theta_{old}$ are the network parameters at the beginning of the policy iteration, $A(s,a)$ is the advantage function, which is estimated using the generalized advantage estimator \citep{schulman2015high}. 
The second term is referred to as value function error term which correspond to learning value function $V_{\theta}(s)$, where $\gamma$ is the discount factor. Finally, the last term $S[\pi_{\theta}]$, corresponds to the entropy of the learned policy, which is added to ensure sufficient exploration. Parameters $\Theta^{ppo}$ refer to the set of the following parameters: $\theta_{old}, A(s,a), \epsilon, c_v, \gamma, V_{\theta}, s', c_e \textup{ and } S[\pi_{\theta}](s)$.
Readers are referred to the Appendix \ref{app: a} for a detailed definition of these parameters.
The approximate multilevel Monte Carlo estimate of $\mathbb{E}_{s,a,r \sim p_{\theta}} [ J (s,a,r; \theta, \Theta^{ppo}) ]$ is defined as
\begin{align}
         \sum_{l=1}^{l=L} \widehat{\mathbb{E}}^{M_l}_{\substack{s^l,a^l,r^l \sim p^l_\theta \\ \tilde{s}^{l-1},\tilde{a}^{l-1},\tilde{r}^{l-1} = p^{l\Rightarrow l-1}_\theta}} \left [ J(s^l,a^l,r^l;\theta, \Theta^{{ppo}^l}) - J(\tilde{s}^{l-1},\tilde{a}^{l-1},\tilde{r}^{l-1};\theta, \tilde{\Theta}^{{ppo}^{l-1}}) \right ],
         \label{eq: ppo_obj_estimate}
\end{align}
where $J(\tilde{s}^0,\tilde{a}^0,\tilde{r}^0;\theta, \tilde{\Theta}^{{ppo}^0}) \doteq 0$ and $M_l$ is the mini-batch size at level $l$. 
The algorithm \ref{alg: ml_ppo} provides an outline for the multilevel PPO algorithm.
\begin{algorithm}
	\caption{Multilevel Proximal Policy Optimization algorithm} 
	\begin{algorithmic}[1]
	\State Input: $\boldsymbol{\mathcal{E}}=\{\mathcal{E}_1, \ldots, \mathcal{E}_L \}, N, \boldsymbol{T}=\{T_1, \ldots, T_L \}, \boldsymbol{M}=\{M_1, \ldots, M_L \}, K$
		\For {$iteration=1,2,\ldots$}
			\For {$actor=1,2,\ldots,N$}
	        \For { level $l=1,2,\ldots, L$}
	           \State $s^l = \psi_{l-1}^l(\tilde{s}_{T_{l-1}}^{l-1})$ \algorithmiccomment{\textbf{if} $l>1$}
	            \For{$t=1,2,\ldots,T_l$}
	                \State $s^l,a^l,r^l \sim p^l_\theta $
	                \State compute $\Theta^{{ppo}^l}$
	                \State $\tilde{s}^{l-1},\tilde{a}^{l-1},\tilde{r}^{l-1} = p^{l\Rightarrow l-1}_\theta $ \algorithmiccomment{\textbf{if} $l>1$}
	                \State compute $\tilde{\Theta}^{{ppo}^{l-1}}$ \algorithmiccomment{\textbf{if} $l>1$}   
	            \EndFor
	        \EndFor
			\EndFor
			\State gather data $\{\{(s^l,a^l,r^l), (\tilde{s}^{l-1},\tilde{a}^{l-1},\tilde{r}^{l-1})\}_{t=1}^{t=NT_l}\}_{l=1}^{l=L}$, from all actors
			\State optimize equation \ref{eq: ppo_obj_estimate}, with $K$ epochs and minibatch size $M_l\leq NT_l$
			\State update policy network parameters $\theta$
		\EndFor
	\end{algorithmic} 
	\label{alg: ml_ppo}
\end{algorithm}
The inputs are the same as those of the classical PPO algorithm, except that multilevel variables are provided as a set of length $L$: environments at each level $\boldsymbol{\mathcal{E}}=\{\mathcal{E}^1,\ldots, \mathcal{E}^L\}$, number of actors $N$, number of steps at each level $\boldsymbol{T}=\{T^1, \ldots, T^L\}$, number of batches at each level $\boldsymbol{M}=\{M^1, \ldots, M^L\}$ (such that $NT^l \leq M^l$ and $T^1/M^1=\cdots=T^L/M^L$) and number of epochs $K$.
Note that if the sets $\boldsymbol{\mathcal{E}}$, $\boldsymbol{T}$ and $\boldsymbol{M}$ consist of a single value, this algorithm is the same as the classical PPO algorithm where the objective function is estimated using the Monte Carlo method.
We implement this algorithm using a standard RL library, stable baselines 3 \citep{stable-baselines3}.
The implementation details are delineated in Appendix \ref{alg:ppo_stable_baselines}.

\subsection{Multilevel PPO analysis methodology} \label{sect: mlppo_analysis}
We present an analysis methodology to compare the Monte Carlo estimate and the standard multilevel Monte Carlo estimate of the PPO objective function.
The analysis methodology is adopted from \cite{giles2015multilevel}, where the strong and weak convergences of the estimates are checked for predefined mean squared error values.
For convenience of demonstration, let us consider the following shorthand notation.
\begin{align*}
          \widehat{\mathbb{E}}^{N} \left[ Y_l \right] \doteq & \widehat{\mathbb{E}}^{N}_{\substack{s^l,a^l,r^l \sim p^l_\theta \\ \tilde{s}^{l-1},\tilde{a}^{l-1},\tilde{r}^{l-1} = p^{l\Rightarrow l-1}_\theta}} \left [ J(s^l,a^l,r^l;\theta, \Theta^{{ppo}^{l}}) - J(\tilde{s}^{l-1},\tilde{a}^{l-1},\tilde{r}^{l-1};\theta, \tilde{\Theta}^{{ppo}^{l-1}}) \right ], \\
          \widehat{\mathbb{E}}^{N} \left[ J_l \right] \doteq & \widehat{\mathbb{E}}^{N}_{s^l,a^l,r^l \sim p^l_\theta } \left [ J(s^l,a^l,r^l;\theta, \Theta^{{ppo}^{l}}) \right ].
\end{align*}
As a result, the multilevel Monte Carlo estimate of $\mathbb{E} [ J_L ]$ is described as
\begin{equation}
    Y=\sum_{l=1}^L \widehat{\mathbb{E}}^{M_l} \left[ Y_l \right].
    \label{eq: mlmc_est}
\end{equation}
The mean squared error ($MSE$) for this estimator is defined as
\begin{align*}
    MSE&=\mathbb{E} [ (Y - \mathbb{E} [ J_L ])^2 ]\\
    &=\mathbb{V}[Y] + (\mathbb{E}[Y] - \mathbb{E} [ J_L ])^2,
\end{align*}
where $\mathbb{V}[Y]$ is the variance of the estimator and $(\mathbb{E}[Y] - \mathbb{E} [ J_L ])^2$ corresponds to the bias of the estimator.
A sufficient condition on $MSE \leq \varepsilon^2$, expands to $\mathbb{V}[Y]=\varepsilon^2/2$ and $(\mathbb{E}[Y] - \mathbb{E} [ J_L ])^2 \leq \varepsilon^2/2$. 
Under assumption $\mathbb{V}[Y]=\varepsilon^2/2$, the optimal number of samples at each level $M_l$ and the corresponding total cost of the estimator $C_{\tiny{\textup{MLMC}}}$ are calculated as
\begin{align}
    M_l &= 2\varepsilon^{-2} \left ( \sum_{l=1}^L V_l C_l \right ) \sqrt{ \frac{V_l}{C_l} }, \label{eq: m_l} \\
    C_{\tiny{\textup{MLMC}}} &= 2\varepsilon^{-2} \left ( \sum_{l=1}^L V_l C_l \right ) \sqrt{ V_l C_l }, \label{eq: c_mlmc}
\end{align}
where $V_l$ corresponds to the variance estimate $\widehat{\mathbb{V}}^{N_{\infty}} \left[ Y_l \right]$ (defined  as $\widehat{\mathbb{E}}^{N_{\infty}} [ Y^2_l ] - \widehat{\mathbb{E}}^{N_{\infty}} [ Y_l ]^2$) for a large number $N_{\infty}$, of samples and $C_l$ is the computational cost of each sample of $Y_l$. 
The weak convergence test $(\mathbb{E}[Y] - \mathbb{E} [ J_L ])^2 \leq \varepsilon^2/2$, is ensured by the following inequality:
\begin{equation}
    \frac{\max_{l\in \{L-2,L-1,L\}} \widehat{\mathbb{E}}^{N_{\infty}} [ Y_l ] }{(2^\alpha - 1)} \leq \frac{\varepsilon}{\sqrt{2}},
    \label{eq: weak_convergence}
\end{equation}
where $\alpha$ is assumed to be a positive coefficient that explains the decay in the values of $\widehat{\mathbb{E}}^{N_{\infty}} [ Y_l ]$ for the chosen levels in the form $\widehat{\mathbb{E}}^{N_{\infty}} [ Y_l ]=c_12^{-\alpha l}$. It is estimated using linear regression on $\widehat{\mathbb{E}}^{N_{\infty}} [ Y_l ]$ values.
Furthermore, the multilevel estimator $Y$ is compared with the Monte Carlo estimate corresponding to the highest level environment $\mathcal{E}^L$ which is computed as
\begin{equation}
    Y_{\tiny{\textup{MC}}} = \widehat{\mathbb{E}}^{M} \left[ J_L \right].
    \label{eq: mc_estimate}
\end{equation}
The number of samples $M$ and the total cost $C_{\tiny{\textup{MC}}}$ of the Monte Carlo estimate corresponding to the variance of the estimate $\varepsilon^2/2$ are calculated as
\begin{align}
    M = 2\varepsilon^{-2}\frac{V}{C}, \label{eq: m_mc} \\
    C_{\tiny{\textup{MC}}} = 2\varepsilon^{-2}V
    \label{eq: c_mc}
\end{align}
where $V$ is the variance estimate $\widehat{\mathbb{V}}^{N_{\infty}} \left[ J_L \right]$ and $C$ is the computational cost of each sample of $J_L$.

The multilevel PPO analysis is performed in parallel with learning at certain predefined intervals of policy iterations. An outline of the analysis is presented in algorithm \ref{alg: ml_ppo_analysis}.
\begin{algorithm}
	\caption{Analysis of multilevel Proximal Policy Optimization algorithm} 
	\begin{algorithmic}[1]
	\State Input: $\boldsymbol{\mathcal{E}}=\{\mathcal{E}_1, \ldots, \mathcal{E}_L \}, \boldsymbol{\varepsilon}=\{\varepsilon_1, \ldots, \varepsilon_n\}, \{C_1,\ldots,C_L\}, N_{\infty}$
	\State generate samples $\{s^L, a^L, r^L\}_{t=1}^{t=N_{\infty}}$ using the environment $\mathcal{E}^L$ (i.e. $s^L, a^L, r^L \sim p^L_{\theta} $)
	\State generate synchronized samples $\{ \{\tilde{s}^l, \tilde{a}^l, \tilde{r}^l\}_{t=1}^{t=N_{\infty}} \}^{l=L-1}_{l=1} $ on sublevels (where $\tilde{s}^l, \tilde{a}^l, \tilde{r}^l = p^{L\Rightarrow l}_{\theta} $)
	\State compute $\widehat{\mathbb{E}}^{N_{\infty}} \left[ Y_l \right]$, $\widehat{\mathbb{E}}^{N_{\infty}} \left[ J_l \right]$, $\widehat{\mathbb{V}}^{N_{\infty}} \left[ Y_l \right]$ and $\widehat{\mathbb{V}}^{N_{\infty}} \left[ J_l \right]$  using generated data
	\For {$\varepsilon \textup{ in } \boldsymbol{\varepsilon}$}
	    \State compute $M_l$ (equation \ref{eq: m_l})
	    \State estimate multilevel Monte Carlo estimate $Y$ (equation \ref{eq: mlmc_est})
	    \State compute total cost $C_{\tiny{\textup{MLMC}}}$ (equation \ref{eq: c_mlmc})
	    \State compute $M$ (equation \ref{eq: m_mc})
	    \State estimate Monte Carlo estimate $Y_{\tiny{\textup{MC}}}$ (equation \ref{eq: mc_estimate})
	    \State compute total cost $C_{\tiny{\textup{MC}}}$ (equation \ref{eq: c_mc})
	    \State check weak convergence (equation \ref{eq: weak_convergence})
	\EndFor
	\end{algorithmic} 
	\label{alg: ml_ppo_analysis}
\end{algorithm}
To obtain accurate estimates of $\widehat{\mathbb{E}}^{N_{\infty}} \left[ Y_l \right]$, $\widehat{\mathbb{E}}^{N_{\infty}} \left[ J_l \right]$, $\widehat{\mathbb{V}}^{N_{\infty}} \left[ Y_l \right]$ and $\widehat{\mathbb{V}}^{N_{\infty}} \left[ J_l \right]$ a high number of samples $N_{\infty}$, is chosen. 
The samples of sequences are rolled out on the finest level $L$ (corresponding to random variable $s^L, a^L, r^L \sim p^L_{\theta} $) and its synchronized samples are created in parallel on sublevels $l \in \{1,\ldots,L-1\}$ (corresponding to random variables $\tilde{s}^l, \tilde{a}^l, \tilde{r}^l = p^{L\Rightarrow l}_{\theta} $). The notations $s^L, a^L, r^L \sim p^L_{\theta}$ and $\tilde{s}^l, \tilde{a}^l, \tilde{r}^l = p^{L\Rightarrow l}_{\theta}$ are delineated in equation \ref{eq: ml_ana_rollout_notation} as
\begin{equation}
\begin{matrix}
s^L,a^L,r^L \sim p^L_{\theta}
\left\{\begin{matrix}
s^L \sim \mu(s^L)\\ 
a^L \sim \pi_{\theta}(a^L|s^L) \\ 
s'^L \sim \mathcal{P}^L(s^L,a^L) \\ 
r^L = \mathcal{R}^l(s^L, a^L, s'^L),
\end{matrix}\right. & 
\tilde{s}^{l},\tilde{a}^{l},\tilde{r}^{l} = p^{L \Rightarrow l}_{\theta}
\left\{\begin{matrix}
\tilde{s}^{l} = \psi^{l}_L(s^L)\\ 
\tilde{a}^L \sim \pi_{\theta}(a^L|s^L) \\ 
\tilde{a}^{l} = \phi^{l}_L(\tilde{a}^L) \\
\tilde{s}'^{l} \sim \mathcal{P}^{l}(\tilde{s}^{l},\tilde{a}^{l}) \\ 
\tilde{r}^{l} = \mathcal{R}^{l}(\tilde{s}^{l}, \tilde{a}^{l}, \tilde{s}'^{l}).
\end{matrix}\right.
\end{matrix}
\label{eq: ml_ana_rollout_notation}
\end{equation}
The Monte Carlo and multilevel Monte Carlo estimates of the objective function of PPO are computed and compared for a set $\boldsymbol{\varepsilon}=\{\varepsilon_1,\ldots,\varepsilon_n\}$, of MSE accuracy values. The computational effectiveness of the multilevel estimator is demonstrated by comparing its total cost $C_{\tiny{\textup{MLMC}}}$ with the corresponding total cost for the Monte Carlo estimate $C_{\tiny{\textup{MC}}}$ for each accuracy value.

\section{Experiments} \label{sect: experiments}
We present two case studies of simulation environments in which the transition between states is governed by the solution of two partial differential equations that describe the incompressible flow of a single phase through porous medium. The stochasticity of the environments is attributed to an uncertain field of permeability. The governing equations for a single phase flow $c$ of clean water, through a porous medium with porosity $\eta$, consist of the continuity equation coupled with the incompressibility condition that are defined as
\begin{equation}
    \begin{matrix}
\eta \frac{dc}{dt} = cq - \nabla\cdot cv; & \nabla \cdot v = q  & \textup{in } \Omega \subset \mathbb{R}^2.
\end{matrix}
\label{eq: gov_eq}
\end{equation}
Flow velocity $v$ and pressure $p$ are related by Darcy's law: $v=-k/\mu \nabla p$, where $k$ is permeability and $\mu$ is viscosity.
Permeability is treated as a stochastic parameter, and its uncertainty is modeled with a predefined probability distribution.
The source and sink are denoted by $q$, where the source corresponds to the injection rate of uncontaminated fluid (clean water) in the domain $\Omega$, and the sink corresponds to the flow rate of the contaminated fluid at the outlet. 


Two environments with distinct parameters and flow scenarios are designed for demonstration of the proposed multilevel PPO algorithm.
For both cases, the parameter values emulate those of the benchmark reservoir simulations presented in SPE-10 model 2 \citep{christie2001tenth}. 
Environments are denoted ResSim-v1 and ResSim-v2 in the rest of the paper (ResSim is a shorthand term for reservoir simulation).

\subsection{ResSim-v1 parameters}
\begin{figure}
    \centering
    \begin{tabular}{c c}
        \subfloat[ResSim-v1] {\resizebox{0.45\columnwidth}{!} {
        \definecolor{red}{rgb}{1,0,0}
\definecolor{blue}{rgb}{0.08,0.4,0.75}
\begin{tikzpicture}[line cap=round,line join=round,>=triangle 45,x=1cm,y=1cm]
\clip(-3,-3) rectangle (4.5,3);

\fill[line width=0.8pt,fill=black,fill opacity=0.18] (-2,0) -- (2,1) -- (2,0) -- (-2,-1) -- cycle;

\draw [line width=0.8pt] (2,2)-- (2,-2);
\draw [line width=0.8pt] (2,-2)-- (-2,-2);
\draw [line width=0.8pt] (-2,-2)-- (-2,2);
\draw [line width=0.8pt] (-2,2)-- (2,2);

\draw [line width=0.4pt] (-2,0)-- (-2.6,0);
\draw [line width=0.4pt] (-2,-1)-- (-2.6,-1);
\draw [line width=0.4pt] (-2,2)-- (-2.6,2);
\draw [line width=0.4pt] (2,1)-- (2.6,1);
\draw [line width=0.4pt] (2,2)-- (3.2,2);
\draw [line width=0.4pt] (2,-2)-- (3.2,-2);
\draw [line width=0.4pt] (-2,-2)-- (-2,-2.7);
\draw [line width=0.4pt] (2,-2)-- (2,-2.7);

\draw [<->,line width=0.4pt] (-2.3,0) -- (-2.3,2);
\draw [<->,line width=0.4pt] (-2.3,0) -- (-2.3,-1);
\draw [<->,line width=0.4pt] (2.3,1) -- (2.3,2);
\draw [<->,line width=0.4pt] (3,-2) -- (3,2);
\draw [<->,line width=0.4pt] (-2,-2.5) -- (2,-2.5);

\draw (-2.3,1) node[anchor=east] {$l_1$};
\draw (-2.3,-0.5) node[anchor=east] {$w$};
\draw (2.3,1.5) node[anchor=west] {$l_2$};
\draw (3,0) node[anchor=west] {$1200$ ft};
\draw (0,-2.5) node[anchor=north] {$1200$ ft};

\draw [fill=blue] (-1.9,1.9) circle (2.5pt);
\draw [fill=blue] (-1.9,1.7) circle (2.5pt);
\draw [fill=blue] (-1.9,1.5) circle (2.5pt);  
\draw [fill=blue] (-1.9,1.3) circle (2.5pt);
\draw [fill=blue] (-1.9,1.1) circle (2.5pt);
\draw [fill=blue] (-1.9,0.9) circle (2.5pt);
\draw [fill=blue] (-1.9,0.7) circle (2.5pt);
\draw [fill=blue] (-1.9,0.3) circle (2.5pt);
\draw [fill=blue] (-1.9,0.1) circle (2.5pt);
\draw [fill=blue] (-1.9,0.5) circle (2.5pt);
\draw [fill=blue] (-1.9,-1.9) circle (2.5pt);
\draw [fill=blue] (-1.9,-1.7) circle (2.5pt);
\draw [fill=blue] (-1.9,-1.5) circle (2.5pt);
\draw [fill=blue] (-1.9,-1.3) circle (2.5pt);
\draw [fill=blue] (-1.9,-1.1) circle (2.5pt);
\draw [fill=blue] (-1.9,-0.9) circle (2.5pt);
\draw [fill=blue] (-1.9,-0.7) circle (2.5pt);
\draw [fill=blue] (-1.9,-0.3) circle (2.5pt);
\draw [fill=blue] (-1.9,-0.1) circle (2.5pt);
\draw [fill=blue] (-1.9,-0.5) circle (2.5pt);
\draw [fill=red] (1.9,1.9) circle (2.5pt);
\draw [fill=red] (1.9,1.7) circle (2.5pt);
\draw [fill=red] (1.9,1.5) circle (2.5pt);
\draw [fill=red] (1.9,1.3) circle (2.5pt);
\draw [fill=red] (1.9,1.1) circle (2.5pt);
\draw [fill=red] (1.9,0.9) circle (2.5pt);
\draw [fill=red] (1.9,0.7) circle (2.5pt);
\draw [fill=red] (1.9,0.3) circle (2.5pt);
\draw [fill=red] (1.9,0.1) circle (2.5pt);
\draw [fill=red] (1.9,0.5) circle (2.5pt);
\draw [fill=red] (1.9,-1.9) circle (2.5pt);
\draw [fill=red] (1.9,-1.7) circle (2.5pt);
\draw [fill=red] (1.9,-1.5) circle (2.5pt);
\draw [fill=red] (1.9,-1.3) circle (2.5pt);
\draw [fill=red] (1.9,-1.1) circle (2.5pt);
\draw [fill=red] (1.9,-0.9) circle (2.5pt);
\draw [fill=red] (1.9,-0.7) circle (2.5pt);
\draw [fill=red] (1.9,-0.3) circle (2.5pt);
\draw [fill=red] (1.9,-0.1) circle (2.5pt);
\draw [fill=red] (1.9,-0.5) circle (2.5pt);
\end{tikzpicture}
        }}  &
        \subfloat[ResSim-v2] {\resizebox{0.45\columnwidth}{!}{
        \definecolor{red}{rgb}{1,0,0}
\definecolor{blue}{rgb}{0.08,0.4,0.75}
\begin{tikzpicture}[line cap=round,line join=round,>=triangle 45,x=1cm,y=1cm]
\clip(-3,-4) rectangle (4,4);

\draw [line width=0.8pt] (1.5,3)-- (1.5,-3);
\draw [line width=0.8pt] (1.5,-3)-- (-1.5,-3);
\draw [line width=0.8pt] (-1.5,-3)-- (-1.5,3);
\draw [line width=0.8pt] (-1.5,3)-- (1.5,3);

\draw [line width=0.4pt] (1.5,3) -- (2,3);
\draw [line width=0.4pt] (1.5,-3) -- (2,-3);
\draw [line width=0.4pt] (1.5,-3) -- (1.5,-3.7);
\draw [line width=0.4pt] (-1.5,-3) -- (-1.5,-3.7);

\draw [<->,line width=0.4pt] (2,3) -- (2,-3);
\draw [<->,line width=0.4pt] (1.5,-3.5) -- (-1.5,-3.5);

\draw (0,-3.6) node[anchor=north] {$620$ ft};
\draw (2,0) node[anchor=west] {$1820$ ft};

\draw [fill=red] (-1.4,2.9) circle (2.5pt);
\draw [fill=red] (-1.4,2) circle (2.5pt);
\draw [fill=red] (-1.4,1) circle (2.5pt);
\draw [fill=red] (-1.4,0) circle (2.5pt);
\draw [fill=red] (-1.4,-1) circle (2.5pt);
\draw [fill=red] (-1.4,-2) circle (2.5pt);
\draw [fill=red] (-1.4,-2.9) circle (2.5pt);

\draw [fill=red] (1.4,2.9) circle (2.5pt);
\draw [fill=red] (1.4,2) circle (2.5pt);
\draw [fill=red] (1.4,1) circle (2.5pt);
\draw [fill=red] (1.4,0) circle (2.5pt);
\draw [fill=red] (1.4,-1) circle (2.5pt);
\draw [fill=red] (1.4,-2) circle (2.5pt);
\draw [fill=red] (1.4,-2.9) circle (2.5pt);

\draw [fill=blue] (0,2.9) circle (2.5pt);
\draw [fill=blue] (0,2) circle (2.5pt);
\draw [fill=blue] (0,1) circle (2.5pt);
\draw [fill=blue] (0,0) circle (2.5pt);
\draw [fill=blue] (0,-1) circle (2.5pt);
\draw [fill=blue] (0,-2) circle (2.5pt);
\draw [fill=blue] (0,-2.9) circle (2.5pt);

\end{tikzpicture}
        }}
    \end{tabular}
    \caption{schematic of the spatial domain $\Omega$}
    \label{fig:domain_schema}
\end{figure}
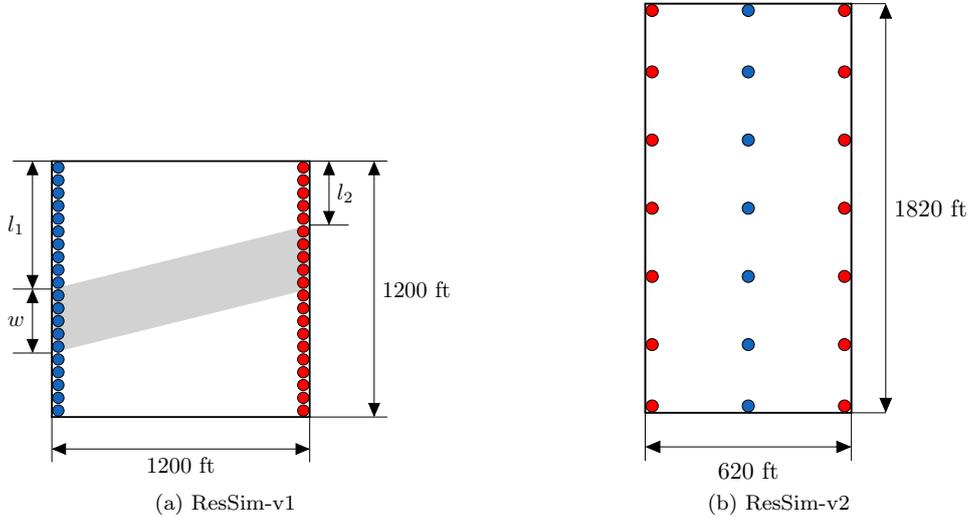

Schematics of the domain $\Omega$ in ResSim-v1 are illustrated in Figure \ref{fig:domain_schema}a.
Viscosity $\mu$ is set to 0.3 cP, while porosity $\eta$ is set to a constant value of 0.2.
According to the convention in geostatistics, the distribution of logarithmic permeability $g=\log{(k)}$ is assumed to be known. 
This logarithmic permeability distribution for test case 1 is inspired by the case study conducted by \citet{brouwer2001recovery}.
In total, 32 injection locations (illustrated with blue circles) and 32 outlet locations (illustrated with red circles) are placed on the left and right edges of the domain, respectively.
The total injection rate is set to a constant value of 2304 $\textup{ft}^2/\textup{day}$.
As illustrated in Figure \ref{fig:domain_schema}a, a linear high-permeability channel (shown in gray) passes from the left to the right side of the domain. 
$l_1$ and $l_2$ represent the distance from the top edge of the domain on the left and right sides, while the width of the channel is indicated by $w$.
These parameters follow uniform distributions defined as $w \sim U(120, 360)$, $l_1 \sim U(0,L-w)$ and $l_2 \sim U(0,L-w)$, where $L$ is the domain length.
To be specific, the logarithmic permeability $g$ at a location $(x,y)$ is formulated as follows:
\begin{equation*}
    g(x, y) = \left\{\begin{matrix}
\log{(245)} & \textup{if} & \frac{l_2-l_1}{L}x+l_1 \leq y \leq \frac{l_2-l_1}{L}x+l_1+w, \\ 
 &  & \\ 
\log{(0.14)} & \ \ \textup{otherwise}, & 
\end{matrix}\right.
\end{equation*}
where $x$ and $y$ are horizontal and vertical distances from the upper left corner of the domain, as illustrated in Figure \ref{fig:domain_schema}a.
The values for permeability in the channel (245 mD) and the rest of the domain (0.14 mD) are inspired from Upperness log-permeability distribution peak values specified in SPE-10 model 2 case.

\subsection{ResSim-v2 parameters}
Figure \ref{fig:domain_schema}b shows the reservoir domain for ResSim-v2. 
It consists of 14 outlets (illustrated with red circles) located symmetrically on the left and right edges (7 on each edge) of the domain and 7 injections (illustrated with blue circles) located at the central vertical axis of the domain.
The total injection rate is set at a constant value of 9072 $\textup{ft}^2/\textup{day}$ while viscosity and porosity are set to the same values as in ResSim-v1.
The uncertainty distribution of the permeability field is considered to be smoother and spatially correlated, and is modeled as a constrained log-normal distribution.
Logarithmic permeability samples are created using ordinary kriging methodology, which are constrained with a constant value of 2.41 logarithmic permeability at injection and outlet locations.
The exponential variogram model used for the kriging is defined as
\begin{equation*}
    \gamma(r) = \sigma^2 \left (1 - \exp{ \left (- \sqrt{ \left(\frac{r_x}{l_x} \right)^2 + \left( \frac{r_y}{l_y} \right)^2 } \right ) } \right)
\end{equation*}
where $r_x$ and $r_y$ are $x$ and $y$ projections of the distance $r$. The variance of the process $\sigma$ is set to 5, while the length scales $l_x$ and $l_y$ are set to 620 ft (width of the domain) and 62 ft (10\% of domain width), respectively. The samples of permeability fields are further rotated clockwise with the angle $\pi/8$. 
In this study, the above-mentioned kriging process is performed using the geostatistics library gstools \citep{sebastian_muller_2019_2541735}.
 
\subsection{Reinforcement learning task} \label{sect: rl_formulation}
In the context of reinforcement learning, the state $s$ is represented by a set of variables $\{c,k,\eta,\mu\}$, while the action $a$ is represented by the source/sink term $q$ at each control step. 
We employ finite-volume discretization of governing equations \ref{eq: gov_eq} as detailed in \citet{aarnes2007introduction} which is treated as a transition function $\mathcal{P}$ between states at time $t_m$ and $t_{m+1}$.
The total time of the simulation is divided into five control steps, which form an episode with finite horizon.
As a result, the task is to learn a policy $\pi_{\theta}(a|s)$ that selects the optimal values of $q$ that maximize the cumulative reward defined as
\begin{equation}
    \sum_{m=1}^{5} \frac{1}{\phi |\Omega|} \int_{t_{m}}^{t_{m-1}} \left ( \int_{\Omega} \min(q,0)(1-c) d\Omega \right ) dt,
    \label{eq: cum_reward}
\end{equation}
where $|\Omega|$ refers to the area of the domain. 
This cumulative reward refers to the sweep efficiency of the injected clean water, which ranges from 0 to 1.
Furthermore, in the context of temporal difference learning, the reward at time $t_m$ is formulated as the term inside the summation operator of equation \ref{eq: cum_reward}.
To represent the stochasticity of the task, a random sample of permeability is chosen from a finite set of permeabilities for each episode in the learning process. This finite set of permeability samples is achieved with a cluster analysis (please refer to Appendix \ref{app: cluster} for the cluster analysis formulation used in this paper). In order to demonstrate application for a partially observable system, the policy network input is replaced with an observation vector instead of the above-defined state. Here, the observation vector corresponds to values of concentration and fluid pressure at injection and outlet locations.
Subsequently, the output of the policy network corresponds to the control vector, which consists of weights (with values ranging between 0.001 to 1) representing flow rates at injection and outlet locations.
Note that with such representation of states, the underlying assumption of the Markov property of the transition function is approximated.
Such a system is referred to as a partially observable Markov decision process (POMDP). 
By the definition of POMDP \citep{spaan2012partially}, the policy requires observations and actions from some sort of history or memory of previous control steps to return the action for a certain control step.
However, for the case studies presented, observation from only the previous control step is sufficient for policy representation.

\begin{figure}
    \centering
    \begin{tabular}{c}
        \subfloat[no policy] {\resizebox{\columnwidth}{!} {
        \includegraphics[trim={2cm 0 2cm 0},clip]{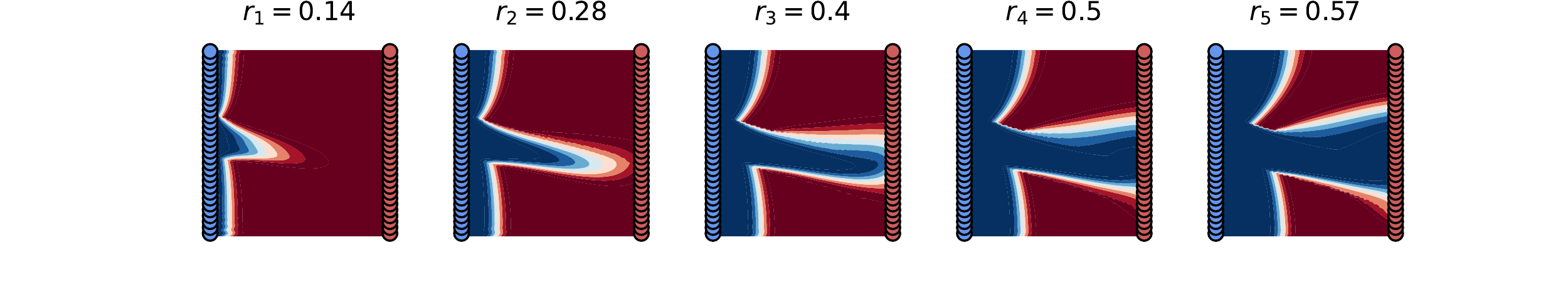}
        }}  \\
        \subfloat[optimal policy] {\resizebox{\columnwidth}{!}{
        \includegraphics[trim={2cm 0 2cm 0},clip]{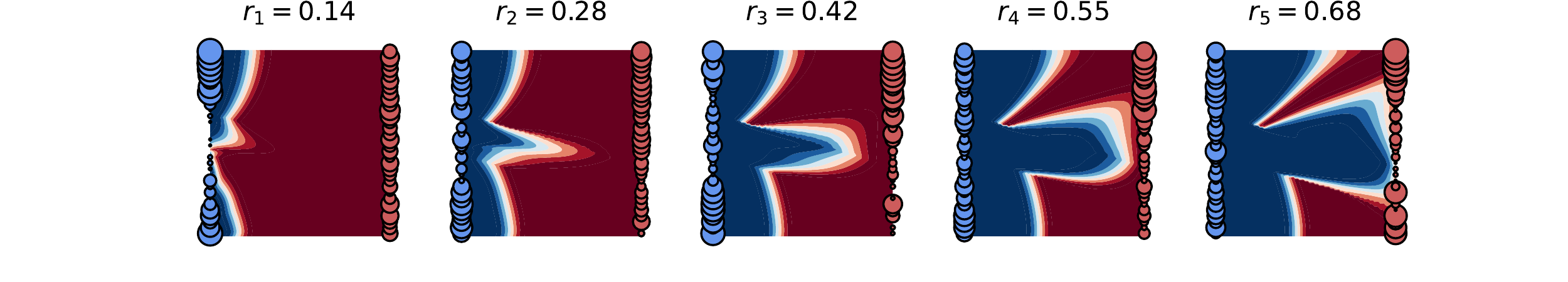}
        }}
    \end{tabular}
    \caption{example policy visualization for ResSim-v1}
    \label{fig: env1_policy}
\end{figure}

\begin{figure}
    \centering
    \begin{tabular}{c}
        \subfloat[no policy] {\resizebox{\columnwidth}{!} {
        \includegraphics{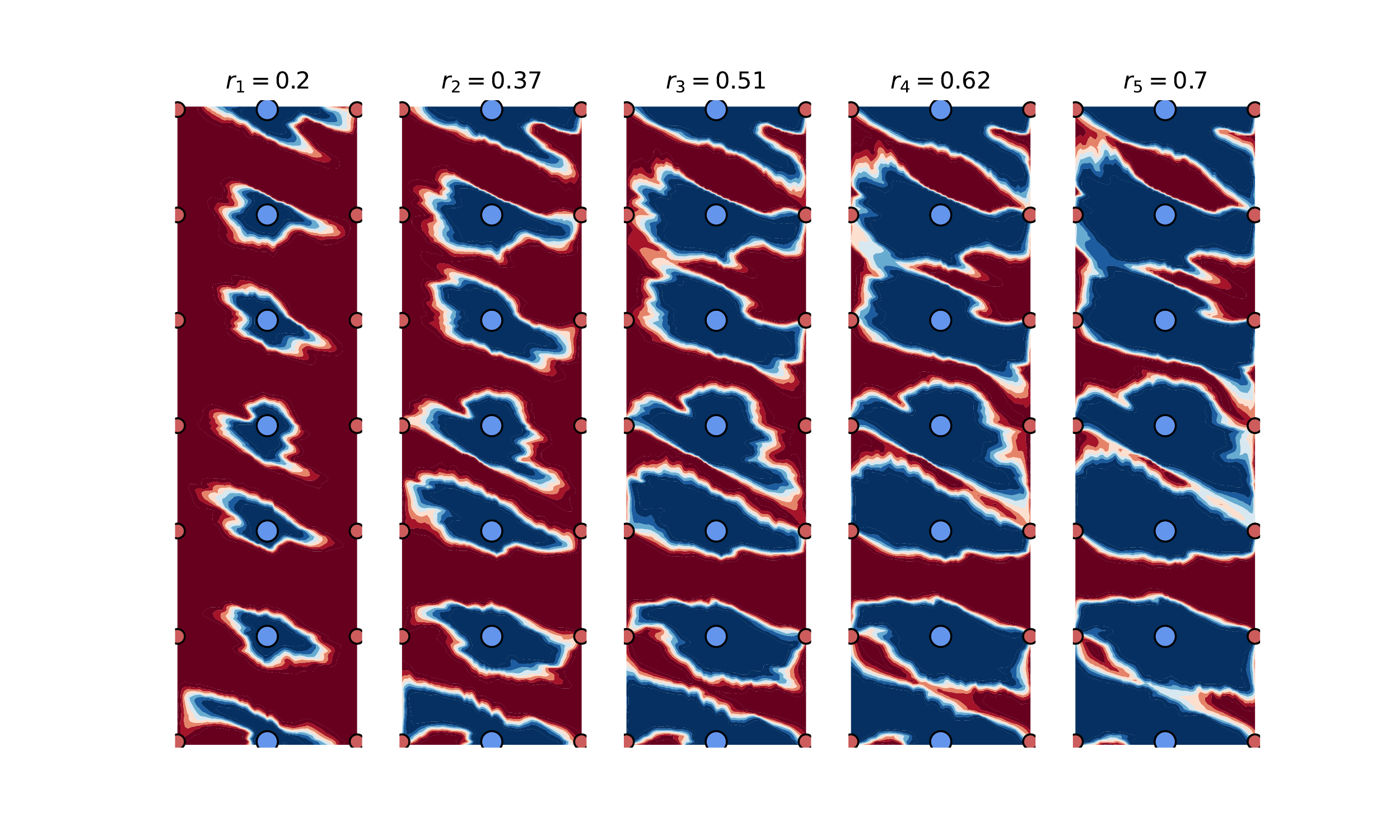}
        }}  \\
        \subfloat[optimal policy] {\resizebox{\columnwidth}{!}{
        \includegraphics{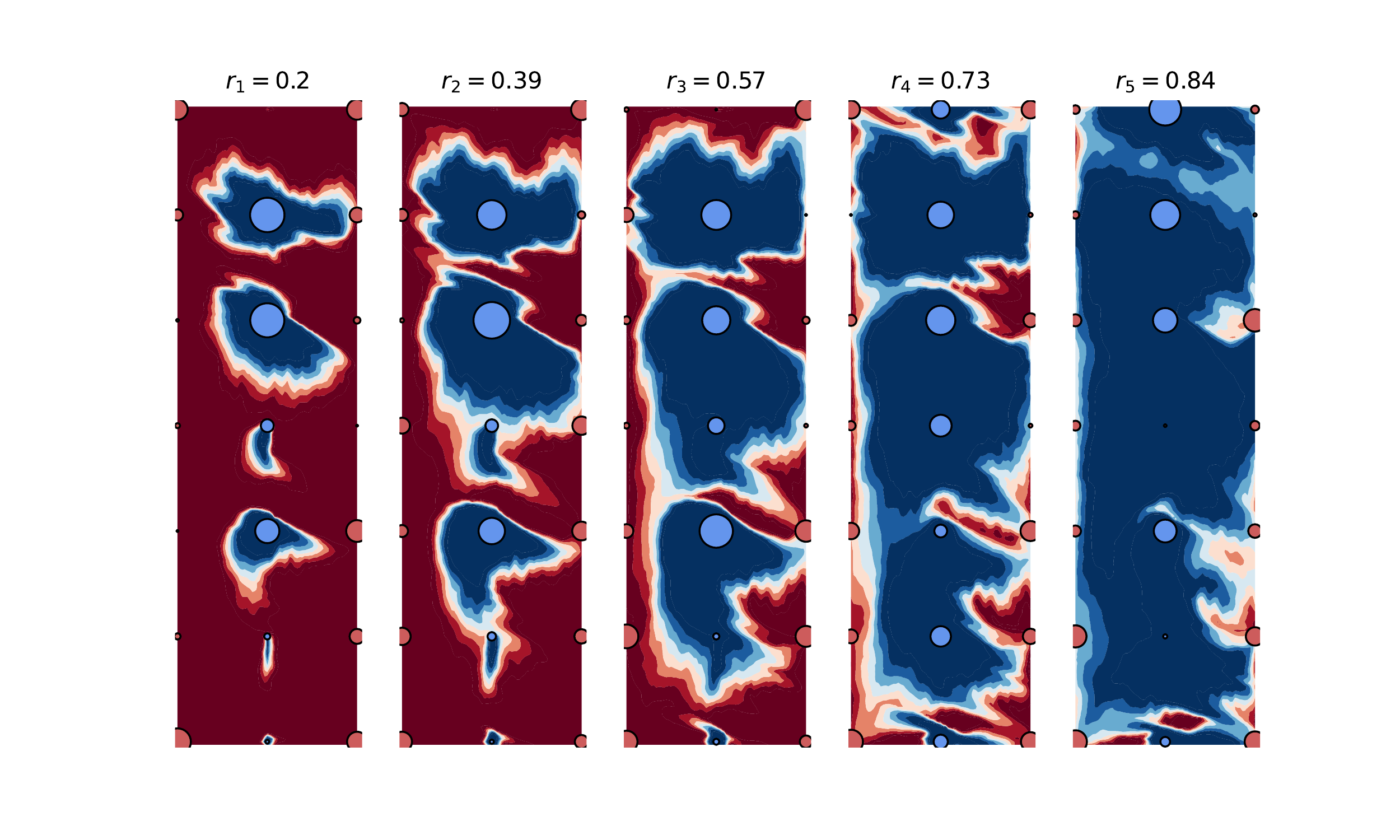}
        }}
    \end{tabular}
    \caption{example of policy visualization for ResSim-v2}
    \label{fig: env2_policy}
\end{figure}

Figure \ref{fig: env1_policy} illustrates visualization of flow through the domain in ResSim-v1.
The injection and outlet locations are indicated with circles in blue and red, respectively, and their radius is proportional to the flow rate.
When ResSim-v1 is operated without a policy (that is, constant injection/outlet rate in all locations), most of the concentration flow takes place in the high-permeability channel, causing poor sweep efficiency in the low-permeability region. 
Figure \ref{fig: env1_policy}a illustrates the flow scenario without a policy for a sample of permeability.
The concentration flow is highlighted with blue in the domain. 
Consequently, the reward which refers to the sweep efficiency corresponds to the ratio of domain area highlighted in blue color to the total domain area.
In other words, optimal policy refers to the choice of actions that increase the domain area in blue (i.e., swept area of the contaminate where the concentration of the clean water is high).
Figure \ref{fig: env1_policy}b illustrates the optimal policy in which the flow through injection / outlet locations near the high permeability area near the channel is restricted. 
As indicated by cumulative rewards at each time, we observe an improvement in total reward at the end of the episode (from 0.57 in no policy to 0.68 with optimal policy).
Similarly, Figure \ref{fig: env2_policy} provides a visualization for the ResSim-v2 environment. 
The optimal policy in this case is to improve the flow rate at locations near the low-permeability locations while restricting the flow rate in locations near the high-permeability region.

\subsection{Multilevel framework formulation}
For multilevel formulation, we consider three levels of environment for the ResSim-v1 environment, where the target task is described by the environment on level 3.
For ResSim-v2 environment, we consider two levels, where the target task is predefined to be at level 2.
The levels for these environments correspond to the grid fidelity of the discretization scheme used to solve the governing equations.
Table \ref{tab: levels} delineates the grid sizes corresponding to each level in the ResSim-v1 and ResSim-v2 environments.
\begin{table}[ht]
    \caption{grid size on each level}
    \centering
    \begin{tabular}{l l l l}
        \hline
         & ResSim-v1 & ResSim-v2\\
        \hline
        level 1 & $32\times32$ & $31\times111$ \\
        level 2 & $64\times64$ & $73\times219$ \\
        level 3 & $128\times128$ & -- \\
        \hline
    \end{tabular}
    \label{tab: levels}
\end{table}
The choice of grid fidelity corresponds to the fact that computational cost and accuracy of model dynamics are proportional to the level.
This is due to the fact that the computational cost and accuracy of a PDE are often proportional to the size of the grid. 
These levels of environment are chosen heuristically for demonstration purpose of the multilevel PPO algorithm.
Although there could be a more systematic approach to choosing these levels, we consider this to be outside the scope of this study.
The state mapping function $\psi_l^{l'}$ maps the state $\{c^l,k^l,\eta^l,\mu^l \}$ for the environment on level $l$ to the state $\{c^{l'},k^{l'},\eta^{l'},\mu^{l'} \}$ for the environment on level $l'$. 
Since porosity $\eta$ and viscosity $\mu$ are set to a constant throughout the domain, we do not need to map them in the function $\psi_l^{l'}$.
As a result, $\psi_l^{l'}$ only maps the concentration $c$ and the permeability $k$ between the level $l$ and $l'$.
When $l$ is larger than $l'$, the mapping occurs from a fine grid to a coarser grid.
This is done by super-positioning a fine grid on a coarse grid and creating coarse partitions on the fine grid. 
The resulting values in each partition are passed through the mean function for concentration values and the harmonic mean for permeability values.
On the contrary, when $l'$ is larger than $l$, the mapping occurs from coarse grid to fine grid.
In this case, the coarse value in each partition is simply assigned to fine grid cells in the corresponding partition.
When it comes to the action mapping function $\phi_l^{l'}$, note that $l'$ is always larger than $l$ for the proposed multilevel framework. As a result, the action $q$ is always mapped from a coarse grid to a finer one.
The coarse to fine mapping is done with the same methodology as $\psi_l^{l'}$, except for the choice of mapping function, which is the sum of the action $q$.
Finally, when $l$ is the same as $l'$, the mapping functions $\psi_l^{l'}$ and $\phi_l^{l'}$ act as an identity function.

\begin{figure}
    \centering
    \begin{tabular}{c c}
        \subfloat[comparison of flow visualization] {\resizebox{0.7\columnwidth}{!} {
        \includegraphics{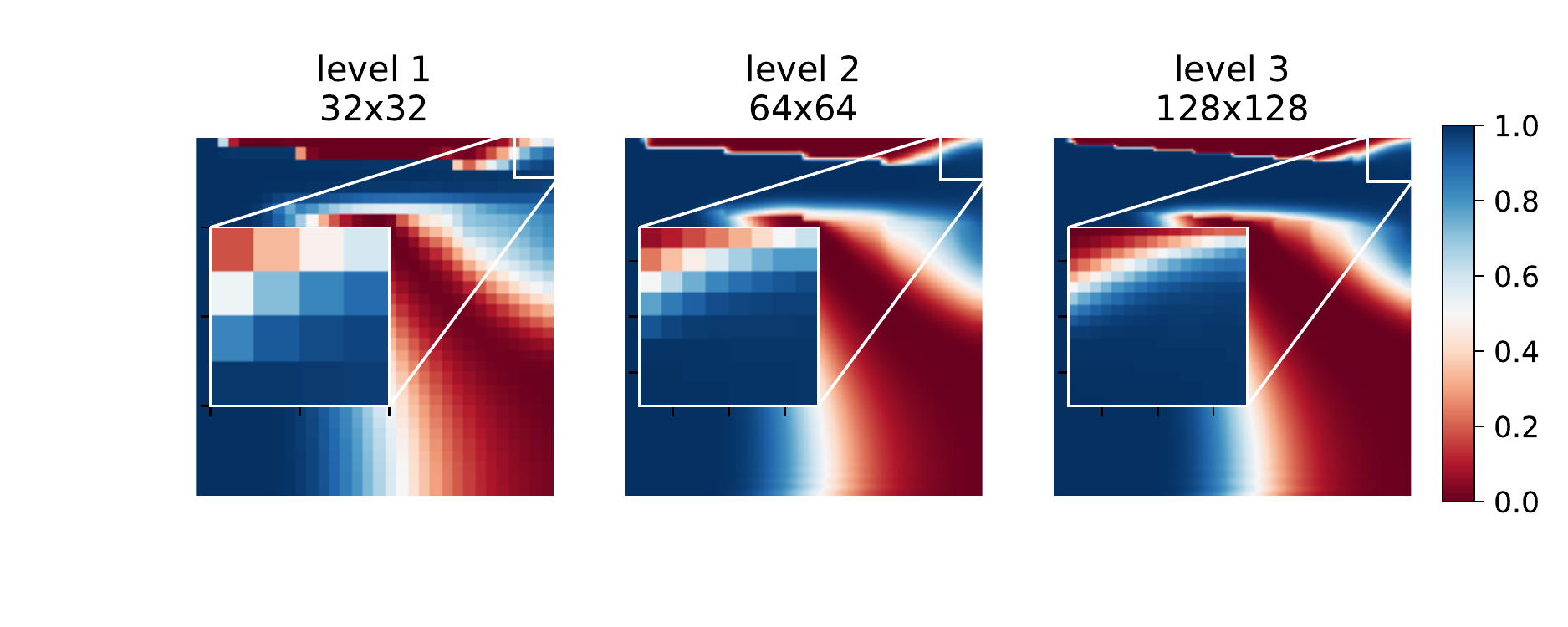}
        }}  &
        \subfloat[computational cost] {\resizebox{0.3\columnwidth}{!}{
        \includegraphics{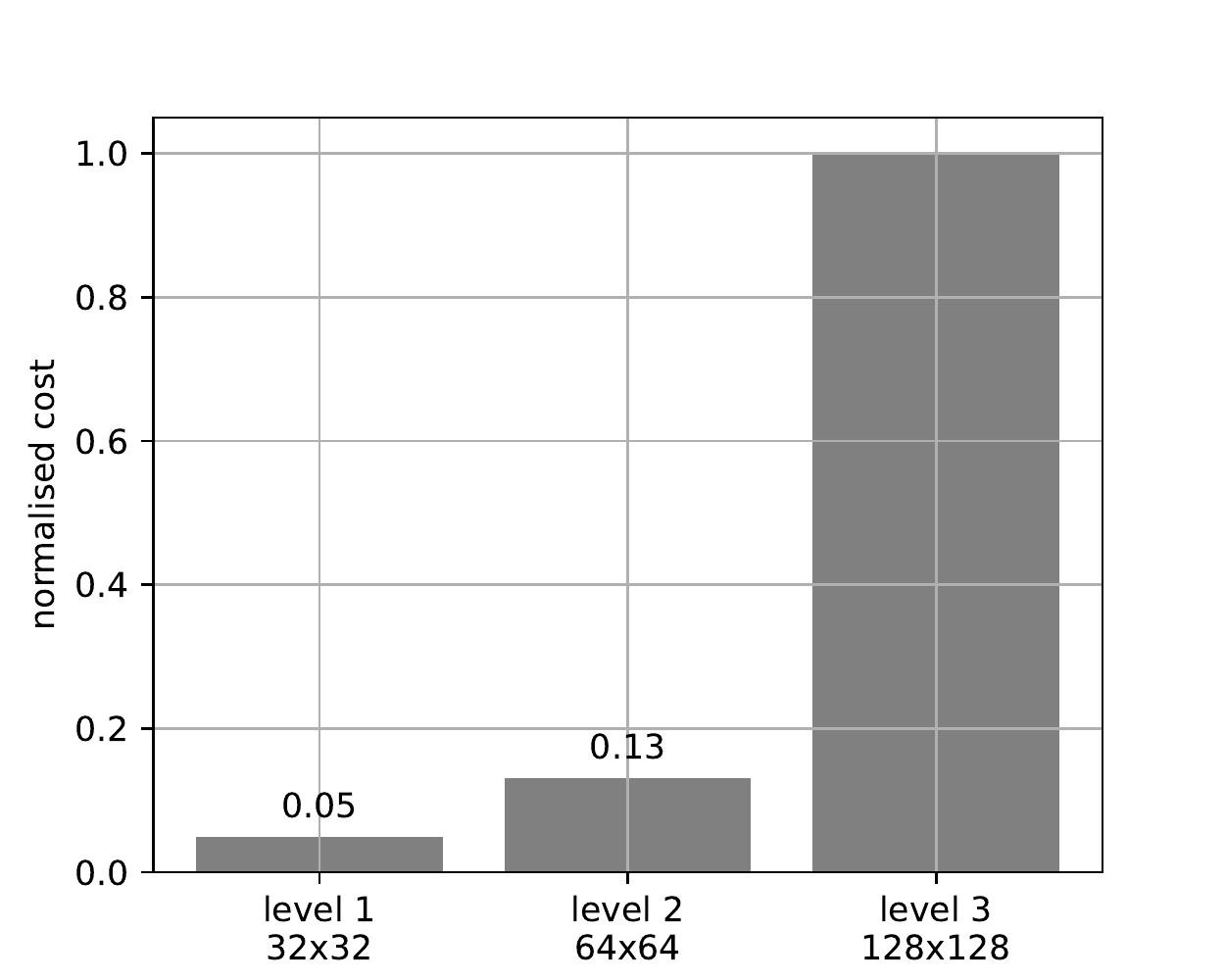}
        }}
    \end{tabular}
    \caption{environment levels for ResSim-v1}
    \label{fig: env1_levels}
\end{figure}

\begin{figure}
    \centering
    \begin{tabular}{c c}
        \subfloat[comparison of flow visualization] {\resizebox{0.4\columnwidth}{!} {
        \includegraphics{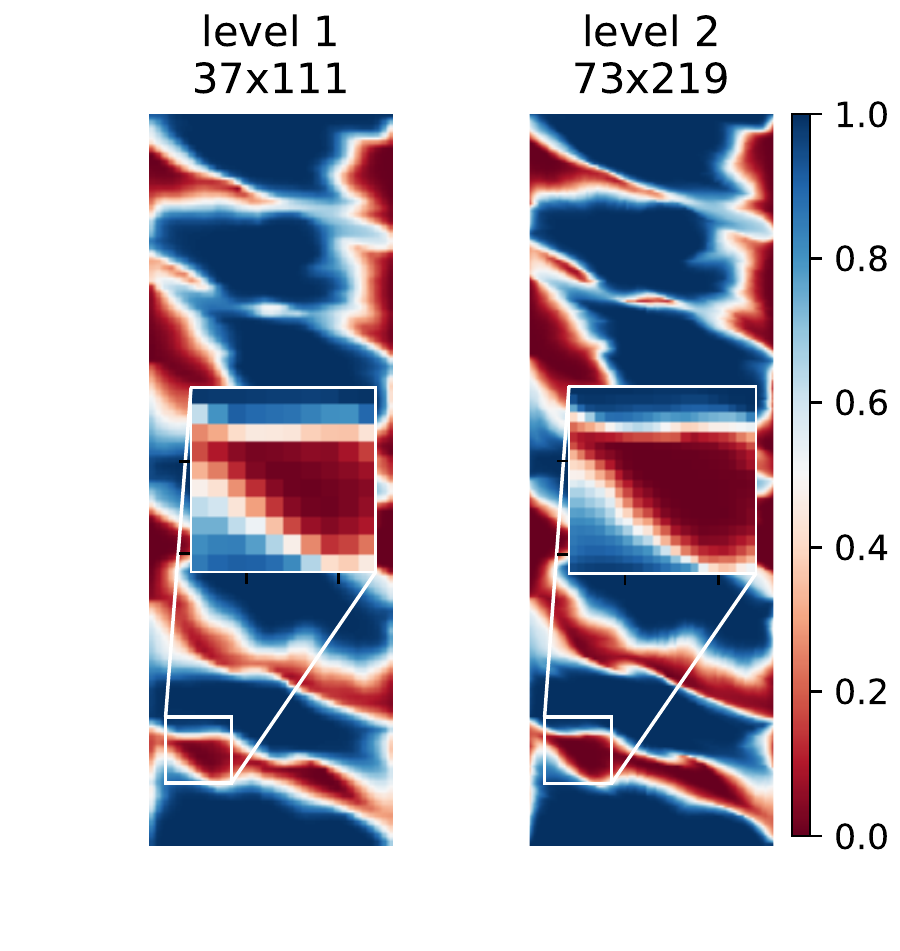}
        }}  &
        \subfloat[computational cost] {\resizebox{0.4\columnwidth}{!}{
        \includegraphics{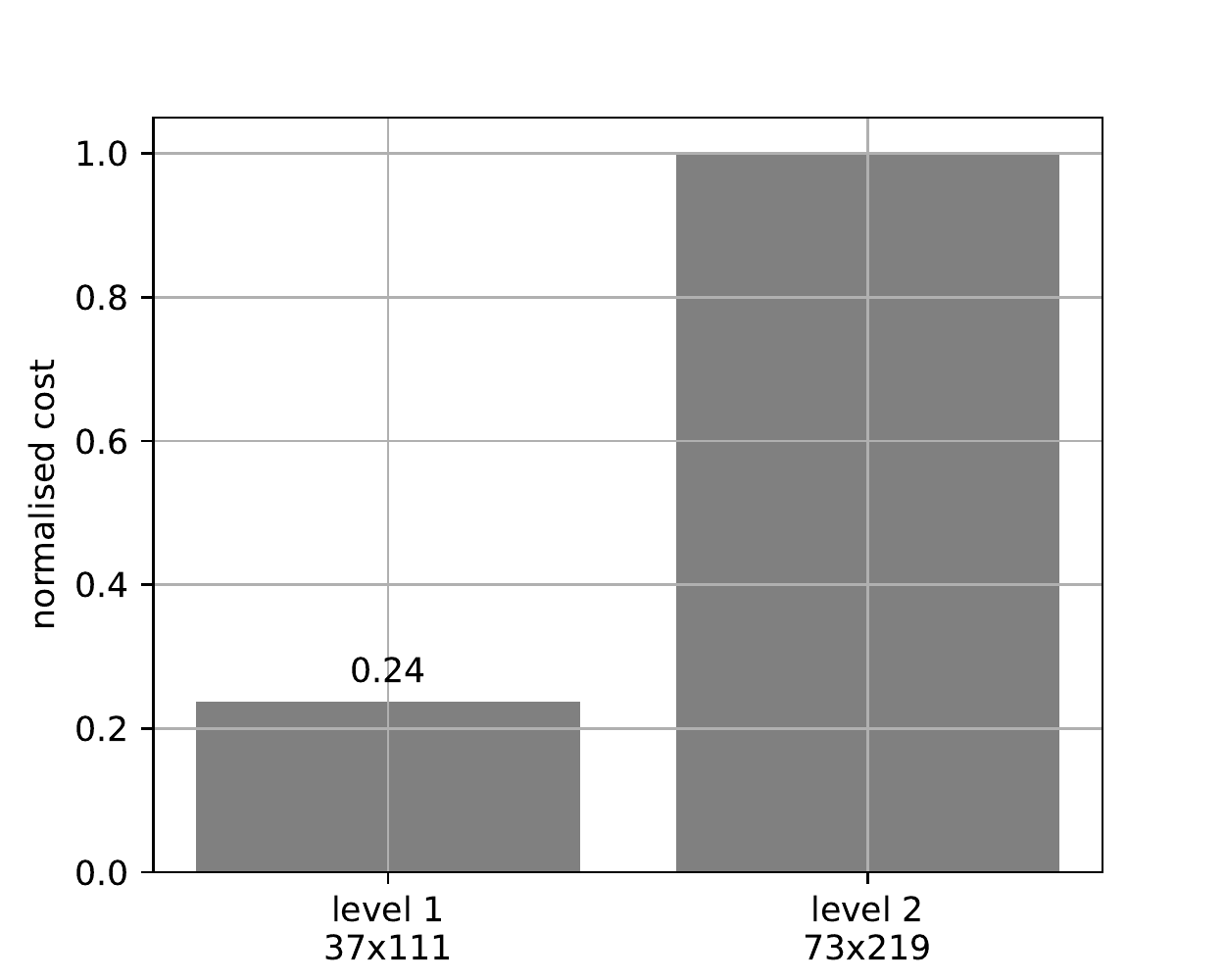}
        }}
    \end{tabular}
    \caption{environment levels for ResSim-v2}
    \label{fig: env2_levels}
\end{figure}

Figure \ref{fig: env1_levels}a illustrates the comparison between flow through the domain at various levels. 
Comparison of the computational cost of the transition function for different levels is illustrated with a bar plot in Figure \ref{fig: env1_levels}b. 
This computational cost is taken as an average value of 100 simulation trials to account for variability.
The computational cost at each level is normalized by dividing it by that corresponding to the target task.
Similar plots for the visualization of two levels of ResSim-v2 are shown in Figure \ref{fig: env2_levels}.

\section{Results} \label{sect: results}
We demonstrate the effectiveness of the multilevel PPO algorithm by comparing its results with the results of the classical single-level PPO algorithm for the target task. 
Table \ref{tab: mlppo_expts} delineates the levels of environments considered in the one-level, two-level, and three-level PPO algorithm (denoted as PPO-1L, PPO-2L, and PPO-3L, respectively).
\begin{table}[ht]
    \caption{levels in each multilevel PPO experiment}
    \centering
    \begin{tabular}{l l l l}
        \hline
         & ResSim-v1 & ResSim-v2\\
        \hline
        PPO-1L & $\{3\}$ & $\{2\}$ \\
        PPO-2L & $\{2,3\}$ & $\{1,2\}$ \\
        PPO-3L & $\{1,2,3\}$ & -- \\
        \hline
    \end{tabular}
    \label{tab: mlppo_expts}
\end{table}
Note that PPO-1L refers to the results of the classical single-level PPO algorithm for the target task.

\subsection{ResSim-v1 results}
\begin{figure}
    \centering
    \includegraphics[width=\columnwidth, trim={5cm 0 5cm 0}, clip]{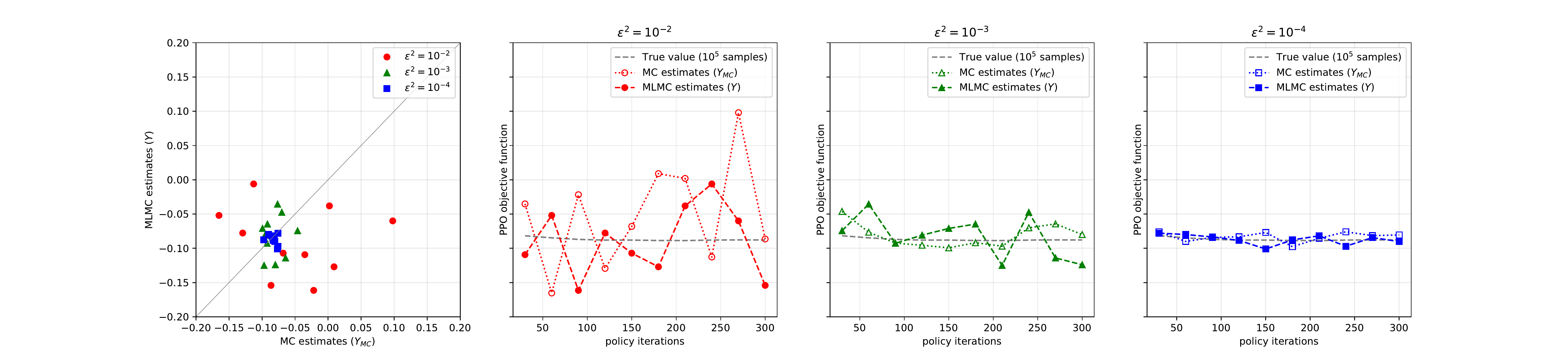}
    \caption{comparison of Monte Carlo and multilevel Monte Carlo estimate of PPO objective function for ResSim-v1}
    \label{fig: case1_mlmc_est}
\end{figure}

\begin{figure}
    \centering
    \begin{tabular}{c c}
        \subfloat[proportions of samples on each level] {\resizebox{0.45\columnwidth}{!} {
        \includegraphics{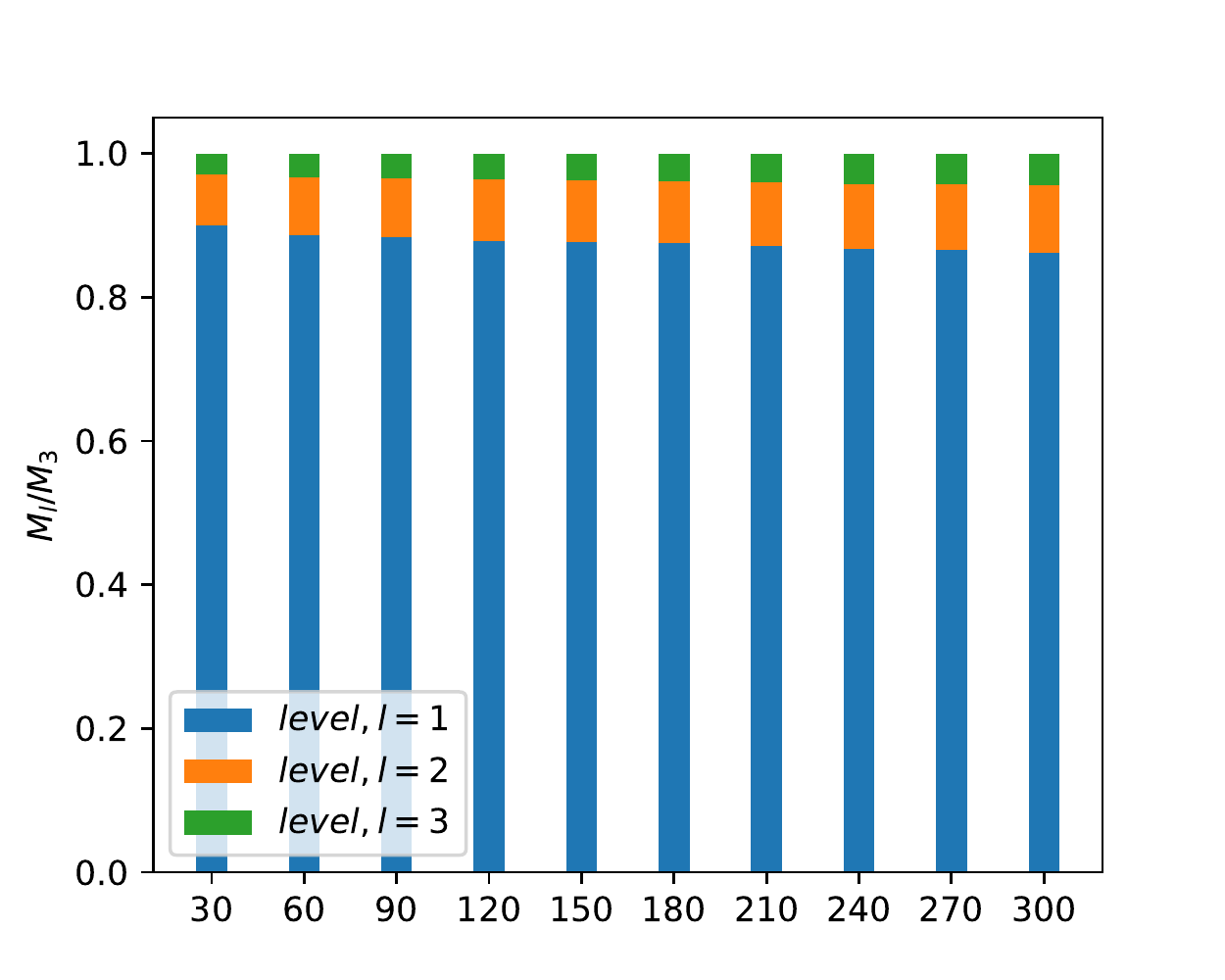}
        }}  &
        \subfloat[comparison of computational cost] {\resizebox{0.45\columnwidth}{!}{
        \includegraphics{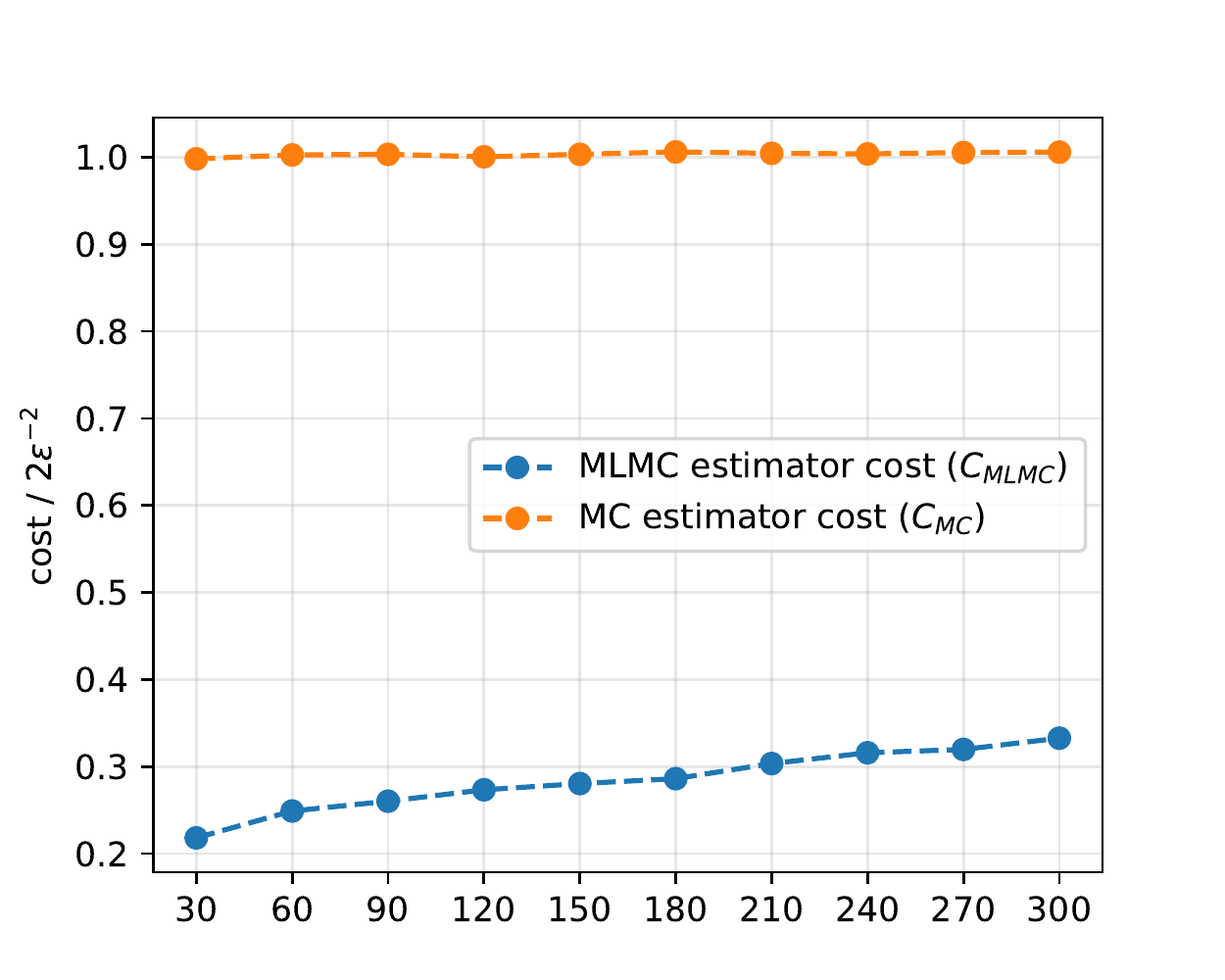}
        }}
    \end{tabular}
    \caption{MLMC analysis results for ResSim-v1}
    \label{fig: case1_mlmc_analysis}
\end{figure}

First, we present the results for multilevel PPO analysis with PPO-1L, which consists of a total of 300 policy iterations. The analysis is performed every 30 iterations. Figure \ref{fig: case1_mlmc_est} illustrates the comparison of Monte Carlo and the three-level Monte Carlo estimate of the objective function with a true value that is estimated using $10^5$ samples (that is, $N_{\infty}$ is set to $10^5$).
The analysis is performed for three values of RMS accuracy: $10^{-2}, 10^{-3}, 10^{-4}$. This is done by setting $\boldsymbol{\varepsilon}=\{\sqrt{10^{-2}},\sqrt{10^{-3}},\sqrt{10^{-4}}\}$ in the analysis.
The cost terms $\{C_1, C_2, C_3\}$, which correspond to the computational cost of each term in the multilevel Monte Carlo estimate, are set to $\{0.1, 0.33, 1.23\}$.
These values are chosen from the computational cost on each level, which are illustrated in Figure \ref{fig: env1_levels}b.
To be specific, $C_1$ refers to the computational cost on level 1 (that is, 0.1). 
The term $C_2$ refers to the computational cost of the difference between synchronized samples at levels 1 and 2, as a result $C_2$ is set as the sum of the computational cost at levels 1 and 2 (i.e., 0.1+0.23).
Similarly, $C_3$ is set as the sum of the computational cost at levels 2 and 3 (that is, 0.23 + 1.0).
As can be seen in figure \ref{fig: case1_mlmc_est}, we see a fairly accurate comparison between Monte Carlo and multilevel Monte Carlo estimates.
Furthermore, these estimates yield more accurate values as we move towards lower values of $\varepsilon^2$. 
This is because the number of samples is inversely related to $\varepsilon^2$ (as stated in equations \ref{eq: m_l} and \ref{eq: m_mc}).
As a result, the number of samples is basically scaled up as we reduce the values of $\varepsilon^2$.
Figure \ref{fig: case1_mlmc_analysis}a illustrates the number of optimal samples at each level of the multilevel estimator.
The numbers of samples are normalized to show the proportions of the samples at each level.
This is done by dividing $M_l$ (from Equation \ref{eq: m_l}) by $M_3$ for all $l \in \{1,2,3\}$.
We see that $M_2$ is approximately $1/10^{th}$ of $M_1$ and $M_3$ is observed to be about half of $M_2$ throughout the learning process.
The comparison between the computational cost of Monte Carlo and the multilevel Monte Carlo estimate is plotted in Figure \ref{fig: case1_mlmc_analysis} b.
Here, the computational cost terms $C_{\textup{MC}}$ and $C_{\textup{MLMC}}$ are divided by $2\varepsilon^{-2}$ to obtain the constant cost terms irrespective of RMS accuracy.
We observe that the computational cost of the multilevel estimate takes only about 20 to 30\% of the Monte Carlo estimate from the analysis.

\begin{figure}
    \centering
    \begin{tabular}{c c}
        \subfloat[learning plot] {\resizebox{0.35\columnwidth}{!} {
        \includegraphics{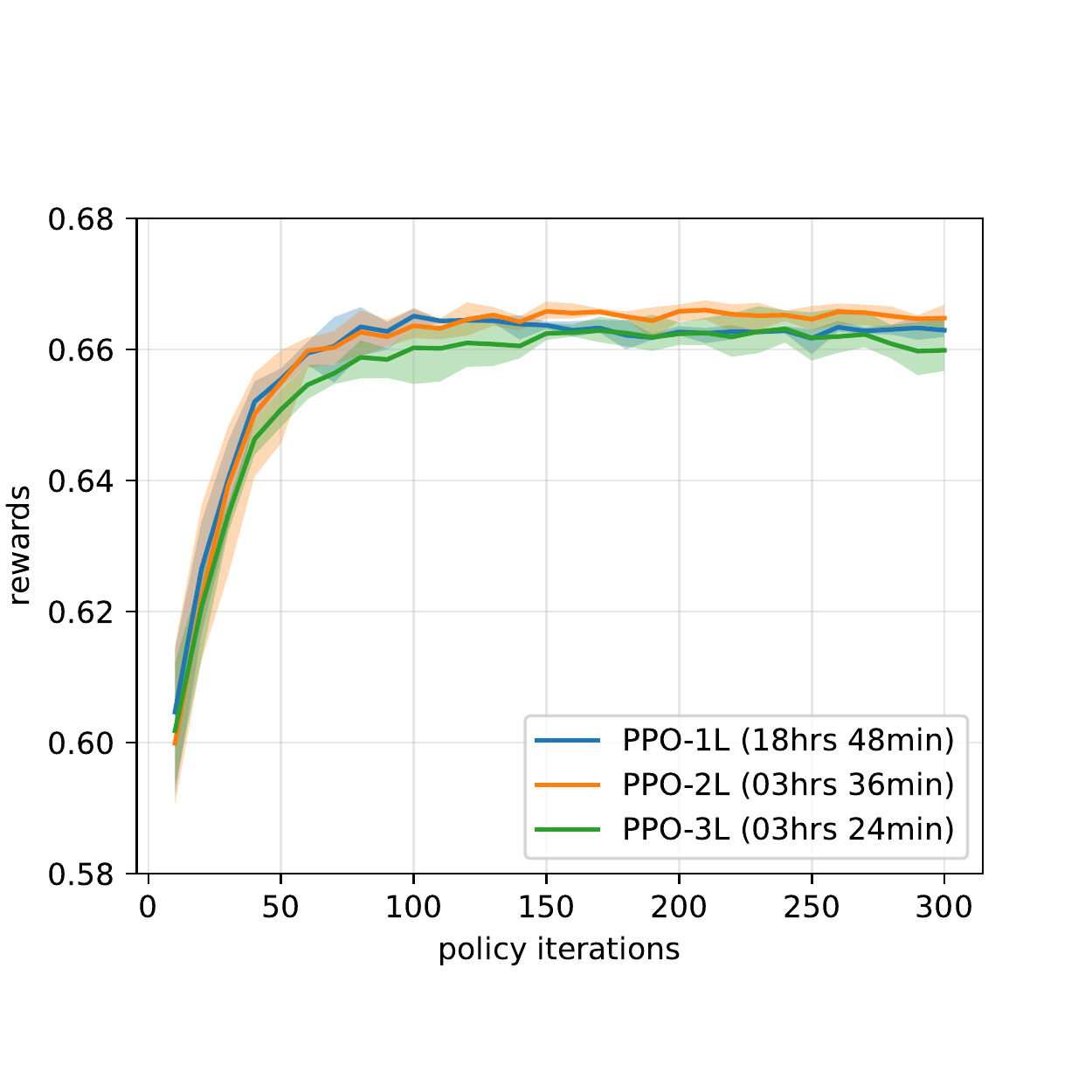}
        }}  &
        \subfloat[policy robustness] {\resizebox{0.65\columnwidth}{!}{
        \includegraphics{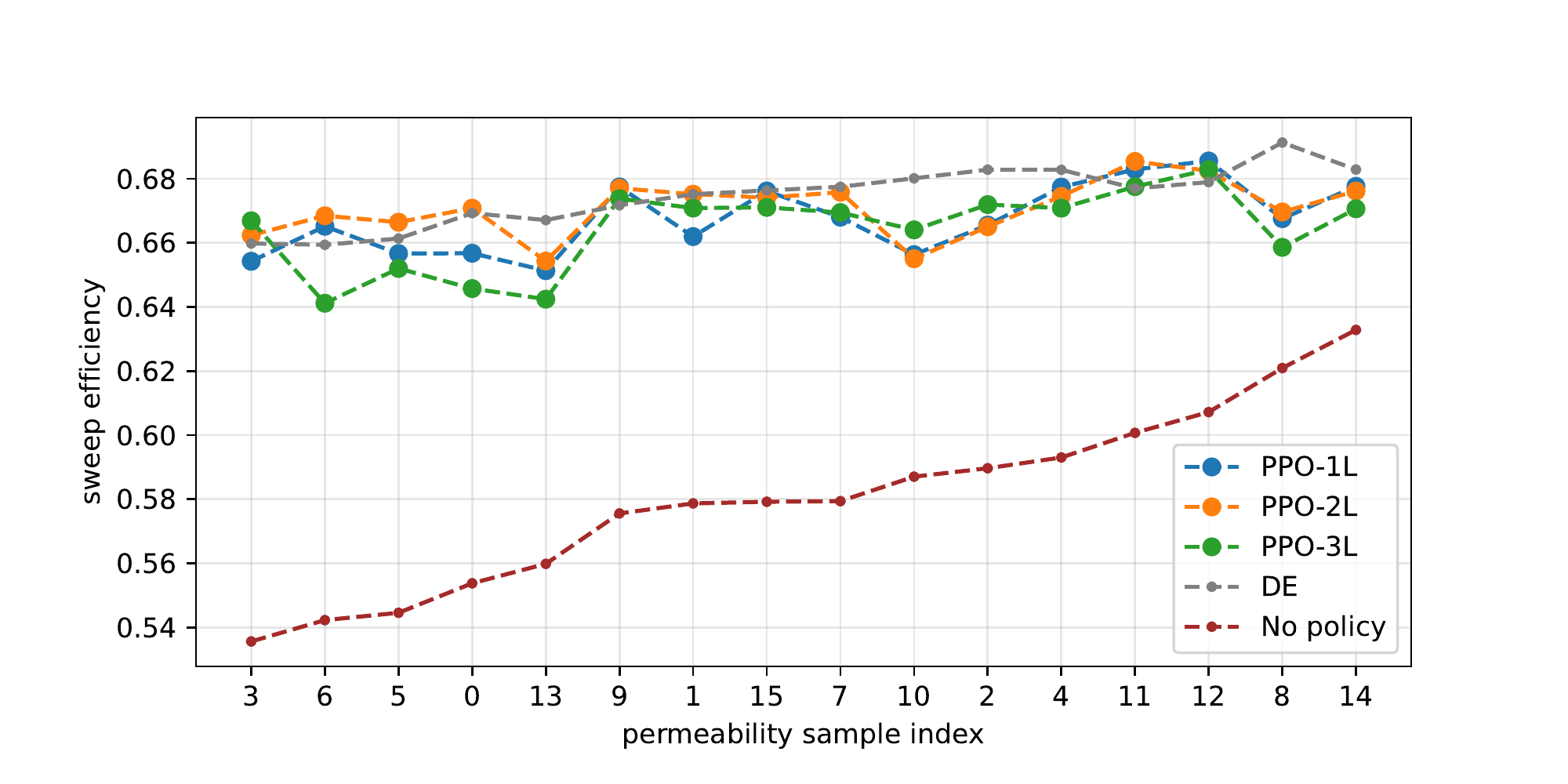}
        }}
    \end{tabular}
    \caption{multilevel PPO results for ResSim-v1}
    \label{fig: env1_results}
\end{figure}

Figure \ref{fig: env1_results} a shows the superposition learning processes for PPO-1L, PPO-2L, and PPO-3L.
The parameters used for the experiments PPO-1L, PPO-2L, and PPO-3L are delineated in the table \ref{tab: case_1_params}.
\begin{table}[ht]
    \caption{parameters of multilevel PPO experiment for ResSim-v1}
    \centering
    \begin{tabular}{c c c c c}
        \hline
         & $\boldsymbol{T}$ & $\boldsymbol{M}$ & $N$ & $K$\\
        \hline
        PPO-1L & $\{50\}$ & $\{250\}$ & 50 & 20 \\
        PPO-2L & $\{70,5\}$ & $\{350,25\}$ & 50 & 20\\
        PPO-3L & $\{80,10,5\}$ & $\{400,50,25\}$ & 50 & 20 \\
        \hline
    \end{tabular}
    \label{tab: case_1_params}
\end{table}
The learning plots are drawn as the average along with the range of values for three distinct seed values.
The parameter $\boldsymbol{M}$, for PPO-1L, PPO-2L, and PPO-3L, is calculated from equation \ref{eq: m_l} for the RMS value $\epsilon^2 = 7.8\times10^{-3}$ where the values of $V_l$ and $C_l$ are taken from the analysis mentioned above.
In other words, we compare the results among PPO-1L, PPO-2L, and PPO-3L for a constant RMS accuracy.
Note that these choices of values are done only in order to demonstrate a fair comparison among PPO-1L, PPO-2L, and PPO-3L.
In practice, it is not required to perform the analysis in order to choose $\boldsymbol{M}$.
Other parameters of the algorithm are tuned to find the convergence for the PPO-3L case first, and these same parameters were used in the PPO-2L and PPO-1L cases.
Figure \ref{fig: env1_results} a shows the evaluation of the environment policy corresponding to the target task.
This policy evaluation is represented with the average reward corresponding to all the permeability samples used in the learning process.
PPO-1L refers to the classical PPO algorithm, which takes around 19 wall clock hours, while PPO-2L and PPO-3L which correspond to the proposed multilevel PPO algorithm achieve the same learning in about three and half hours.
In other words, we save around 82\% computational costs with the proposed algorithm compared to its classical counterpart.
Figure \ref{fig: env1_results}b shows the robustness of the learned policies against uncertainty in permeability.
This is done by plotting rewards for 16 random permeability samples of the uncertainty distribution that were unseen during the learning process.
These results were compared with the optimal solutions obtained using the differential evolution algorithm (implemented using the SciPy library, \citep{2020SciPy-NMeth}), which are denoted DE in Figure \ref{fig: env1_results} b. The algorithm parameters for the PPO and DE algorithms, in this study, are delineated in the Appendix \ref{app: rl_params}.

\subsection{ResSim-v2 results}

\begin{figure}
    \centering
    \includegraphics[width=\columnwidth, trim={5cm 0 5cm 0},clip]{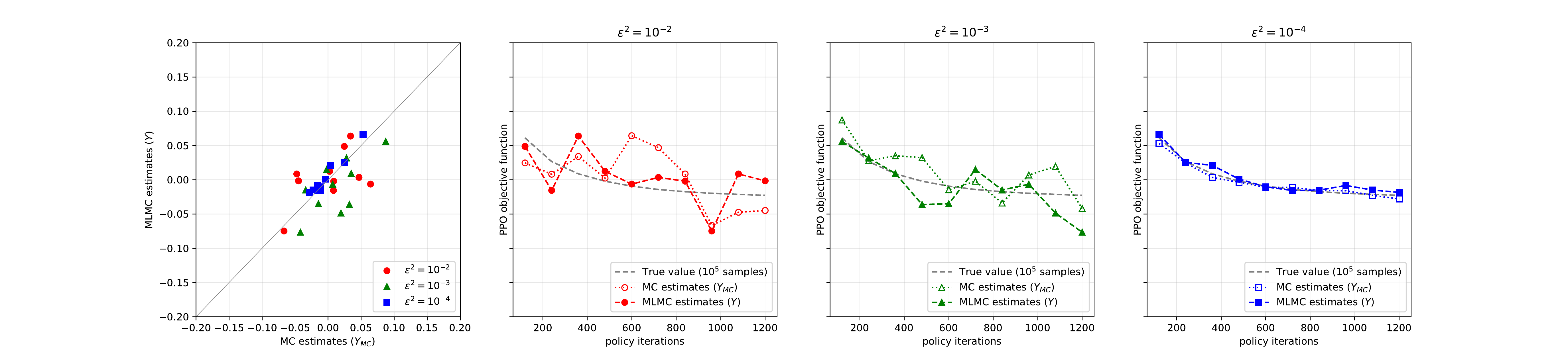}
    \caption{comparison of Monte Carlo and multilevel Monte Carlo estimate of PPO objective function for ResSim-v2}
    \label{fig: case2_mlmc_est}
\end{figure}

\begin{figure}
    \centering
    \begin{tabular}{c c}
        \subfloat[proportions of samples on each level] {\resizebox{0.45\columnwidth}{!} {
        \includegraphics{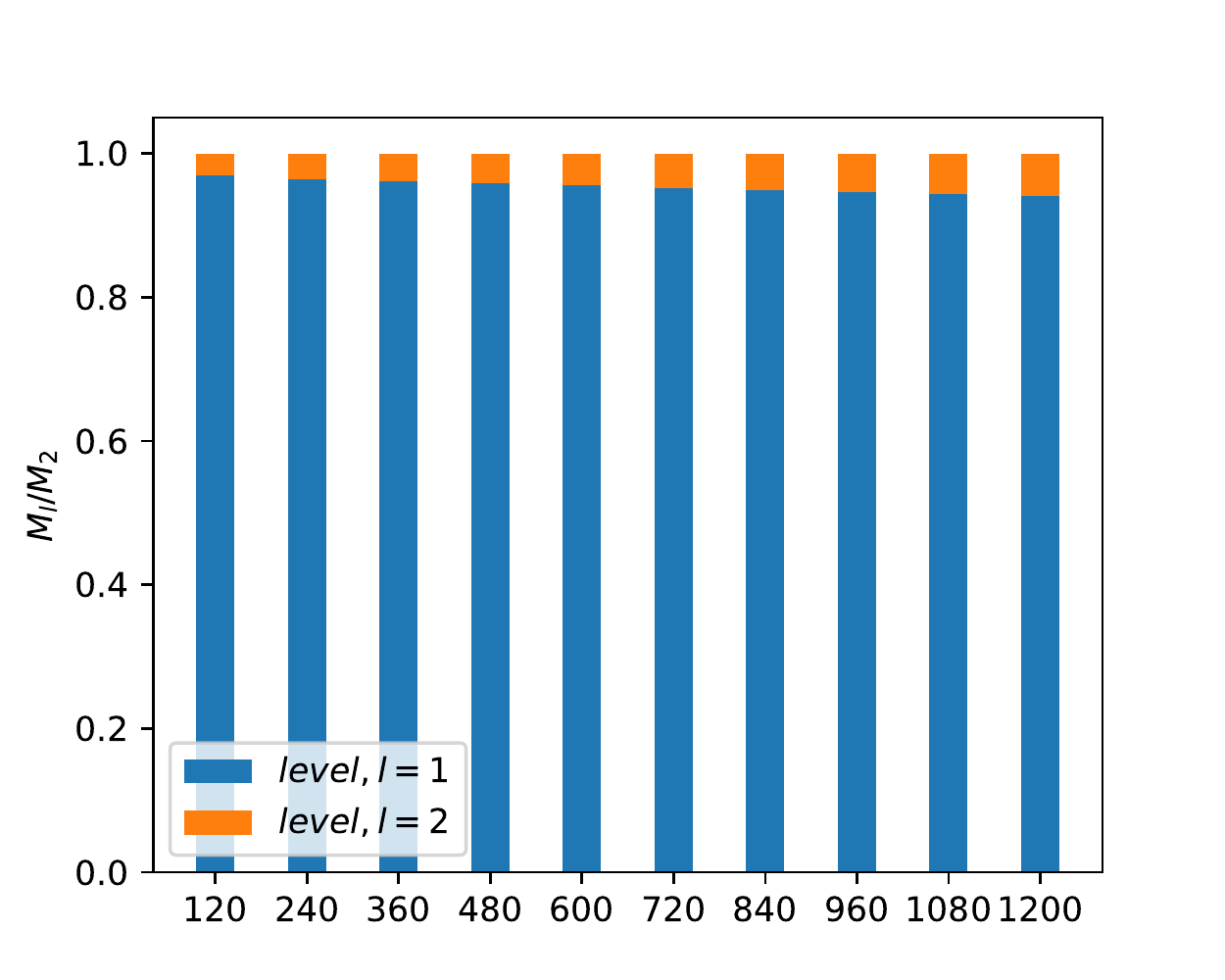}
        }}  &
        \subfloat[comparison of computational cost] {\resizebox{0.45\columnwidth}{!}{
        \includegraphics{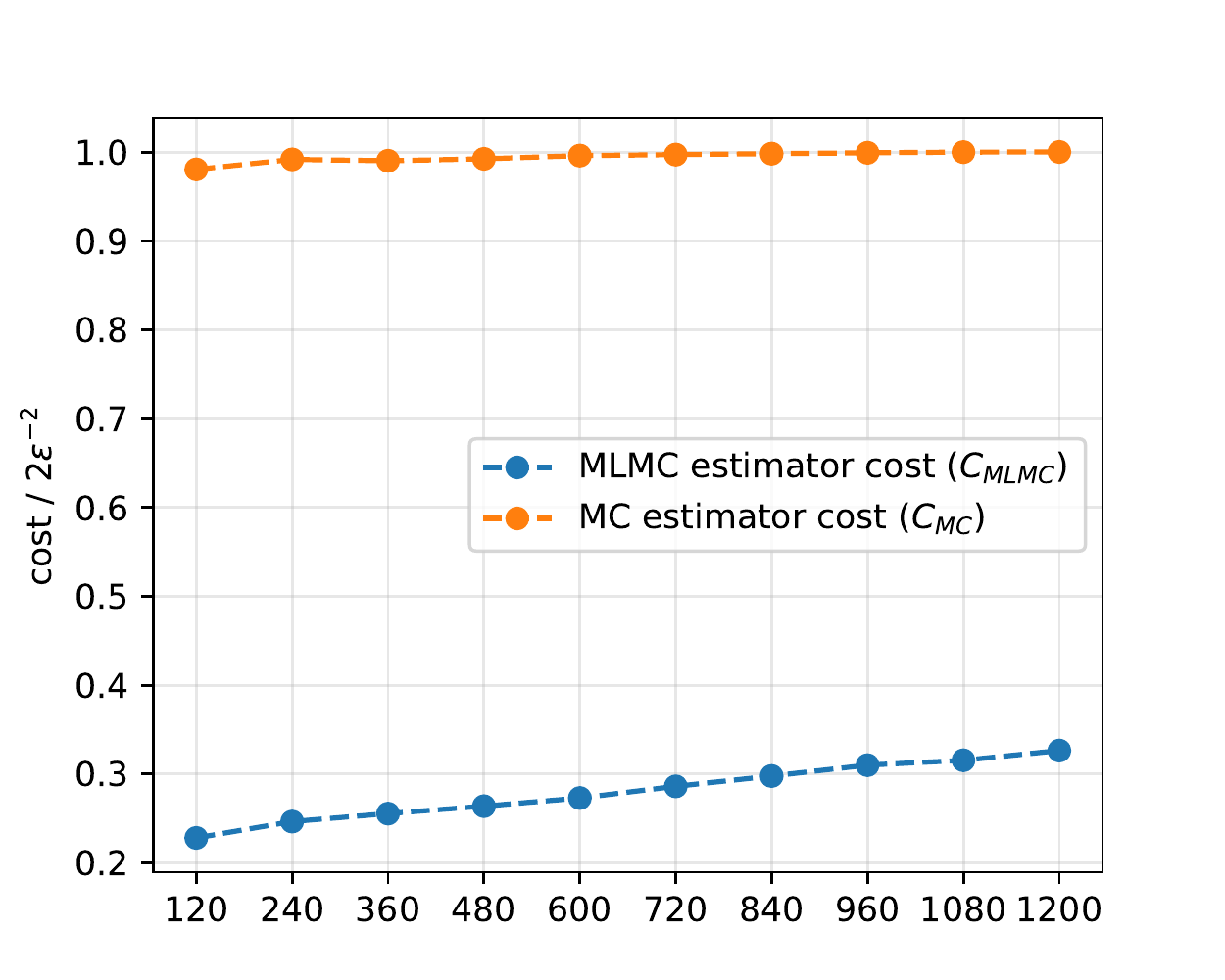}
        }}
    \end{tabular}
    \caption{MLMC analysis results for ResSim-v2}
    \label{fig: case2_mlmc_analysis}
\end{figure}

Similarly to ResSim-v1, we perform multilevel PPO analysis with PPO-1L, which consists of a total of 1200 policy iterations and is performed every 120 iterations. Figure \ref{fig: case2_mlmc_est} illustrates the correlation between Monte Carlo and the multilevel Monte Carlo estimate of the objective function.
The RMS values in $\boldsymbol{\varepsilon}$ are set to $\{\sqrt{10^{-2}},\sqrt{10^{-3}},\sqrt{10^{-4}}\}$ in the analysis.
The cost terms $\{C_1, C_2\}$, which correspond to the computational cost of each term in the multilevel Monte Carlo estimate, are set to $\{0.24, 1.24\}$.
As illustrated in Figure \ref{fig: case2_mlmc_est}, we observe a higher correlation between Monte Carlo and multilevel Monte Carlo estimates for higher RMS accuracy values.
Figure \ref{fig: case2_mlmc_analysis} a illustrates the proportions of the optimal number of samples at both levels of the multilevel estimator.
The comparison between the computational cost of Monte Carlo and the multilevel Monte Carlo estimate is plotted in figure \ref{fig: case2_mlmc_analysis}b.
We observe that the computational cost of the multilevel estimate takes only around 25 to 35\% of the Monte Carlo estimate from the analysis.

\begin{figure}
    \centering
    \begin{tabular}{c c}
        \subfloat[learning plot] {\resizebox{0.35\columnwidth}{!} {
        \includegraphics{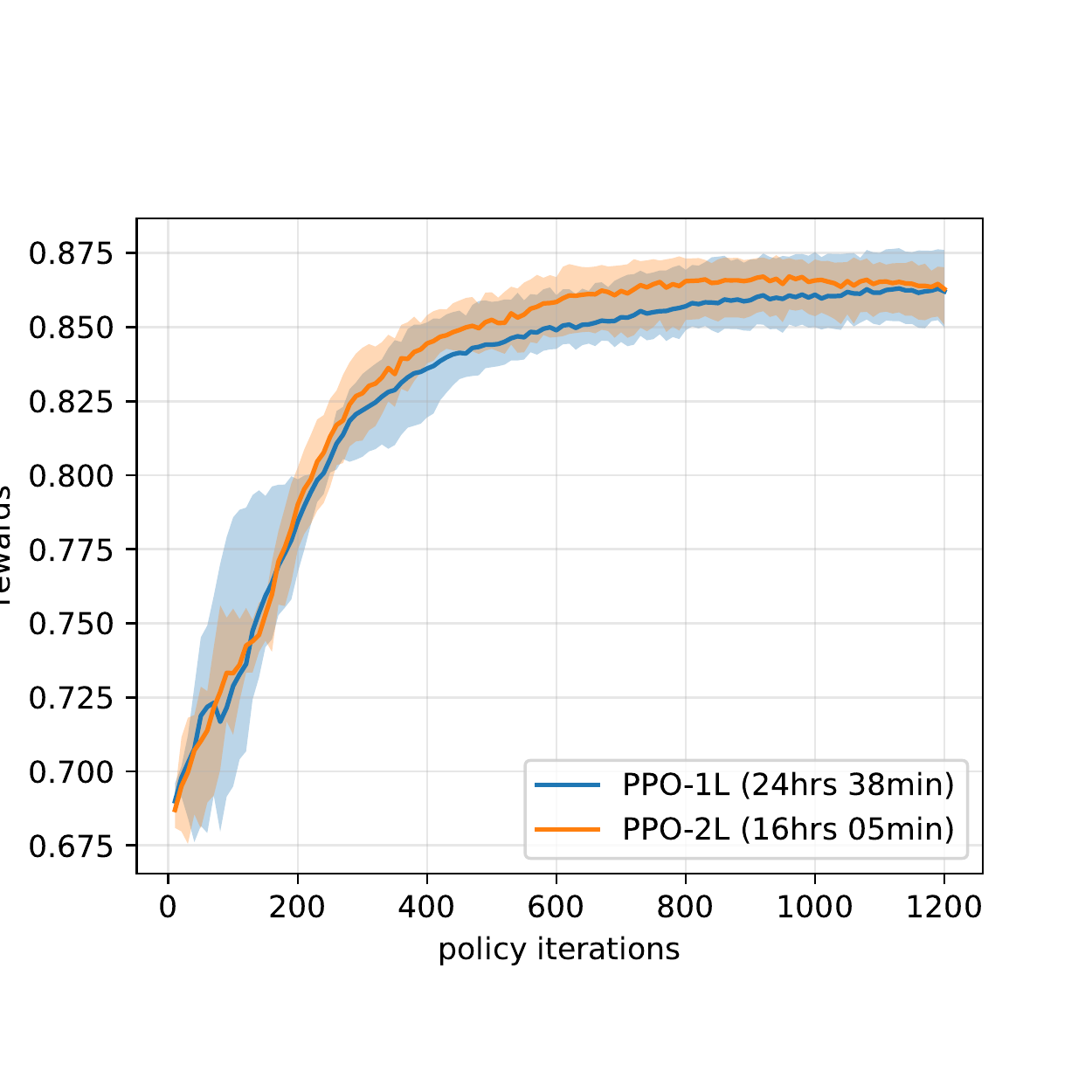}
        }}  &
        \subfloat[policy robustness] {\resizebox{0.65\columnwidth}{!}{
        \includegraphics{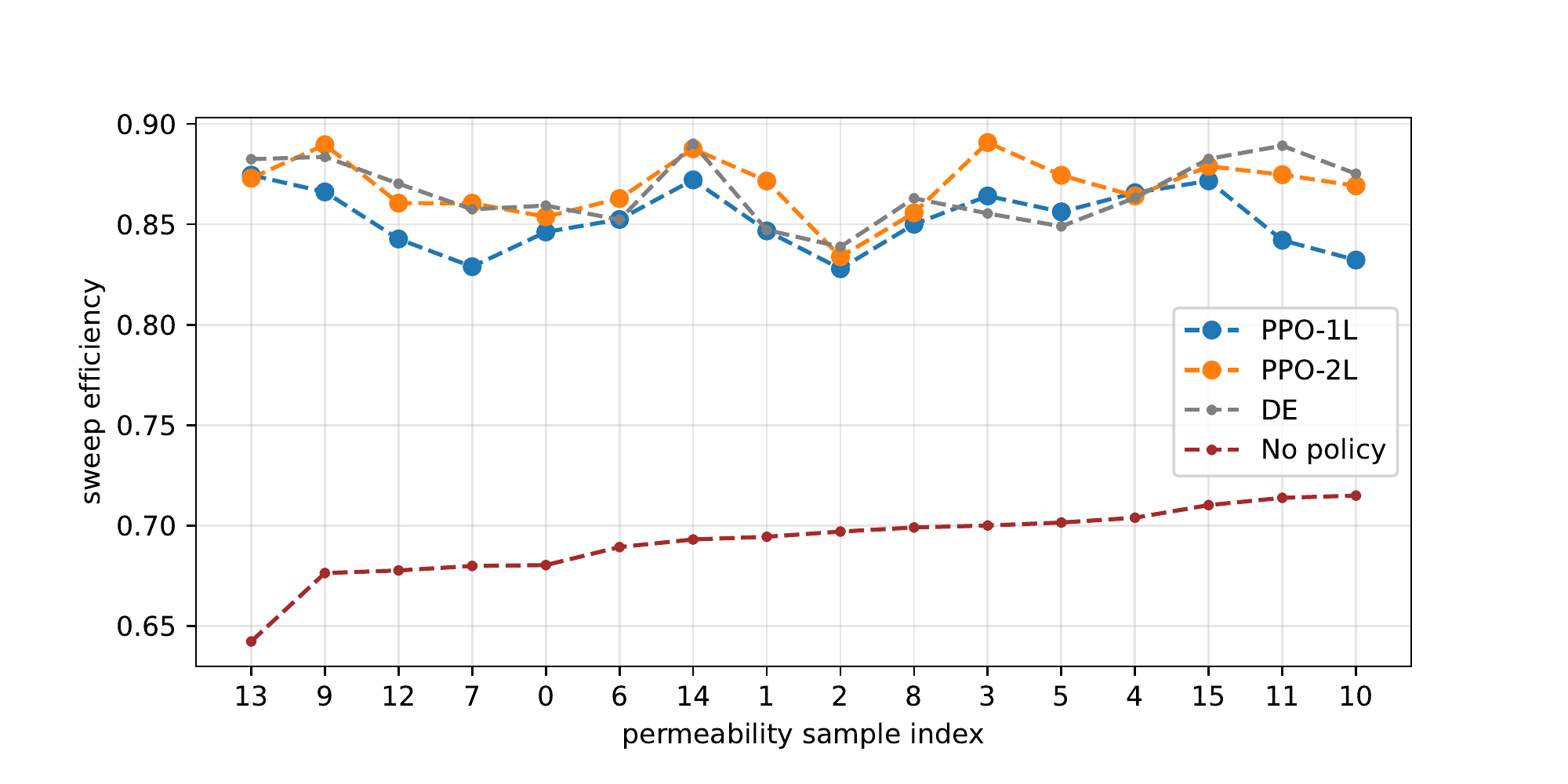}
        }}
    \end{tabular}
    \caption{multilevel PPO results for ResSim-v2}
    \label{fig: env2_results}
\end{figure}

Figure \ref{fig: env2_results} a shows the comparison between the learning processes for PPO-1L and PPO-2L.
The parameters used for the PPO-1L and PPO-2L experiments are delineated in the table \ref{tab: case_2_params}.
\begin{table}[ht]
    \caption{parameters of multilevel PPO experiment for ResSim-v1}
    \centering
    \begin{tabular}{c c c c c}
        \hline
         & $\boldsymbol{T}$ & $\boldsymbol{M}$ & $N$ & $K$\\
        \hline
        PPO-1L & $\{100\}$ & $\{500\}$ & 50 & 20 \\
        PPO-2L & $\{140,15\}$ & $\{700,75\}$ & 50 & 20\\
        \hline
    \end{tabular}
    \label{tab: case_2_params}
\end{table}
In this case, the comparison between PPO-1L and PPO-2L is made for the RMS accuracy value $\epsilon^2 = 3.9\times10^{-3}$.
Similarly to the ResSim-v1 case, the hyperparameters of the algorithm are tuned to find convergence for the PPO-2L case, and the same parameters were used in the PPO-1L case.
Figure \ref{fig: env1_results} a shows the average evaluation of the environment policy corresponding to the target task (level 2).
PPO-1L refers to the classical PPO algorithm which takes around 24 wall clock hours, while PPO-2L, which corresponds to the proposed multilevel PPO algorithm, achieves the same learning in about 16 hours.
In other words, we save around 35\% computational costs with the proposed algorithm compared to its classical counterpart.
Figure \ref{fig: env2_results}b shows the robustness of the learned policies against uncertainty in permeability.

\subsection{challenges and further research direction}
Albeit successful results in learning and analysis of the proposed framework, we believe that this study deserves deeper mathematical investigation and analysis.
In particular, we would like to study and analyze the effect of the approximation introduced in the MLMC estimation. As an introduction to the proposed framework, the experiments presented for multilevel PPO were performed for a specific PDE-based control problem for flow through porous media.
Subsequently, we would like to provide a thorough study with a variety of experiments with the proposed multilevel PPO algorithm.
This study would be mainly aimed at general benchmark problems in which RL is utilized to achieve superhuman controls, but with excessive computational costs. 
Furthermore, while tuning the algorithm parameters for the multilevel PPO algorithm, we observed that increasing the number of levels, the learning rate, and the clip range had an adverse effect on the learning convergence. 


\section{Conclusions} \label{sect: conclusion}
A multilevel framework for deep reinforcement learning is introduced in which the learned agent interacts with multiple levels of PDE-based environments where the level corresponds to the grid fidelity of PDE discretization.
We present a mathematical framework that allows the synchronized implementation of task trajectories at multiple environmental levels.
The presented approximate MLMC estimate is at the heart of the proposed multilevel framework.
We also present a novel multilevel variant of the classical PPO algorithm based on the proposed multilevel framework.
The computational efficiency of this multilevel PPO algorithm is illustrated for two environments for which model dynamics is represented with PDEs describing an incompressible single-phase fluid flow through a porous medium. 
We observe substantial computational savings in the case studies presented (approximately 82\% and 35\%, respectively).

As a future scope of this study, we aim to analyze the effect of the presented approximation to standard MLMC estimation. For the multilevel PPO algorithm, this can be done by extending the analysis methodology (presented in Section \ref{sect: mlppo_analysis}) for the approximate MLMC estimate. We also aim to provide a future study to benchmark multilevel PPO algorithm performance on a variety of environments.  

\clearpage
\appendix

\section{Examples of objective functions for different deep RL algorithms} \label{app: a}
Examples of the objective function $\mathbb{E}_{s,a,r \sim p_{\theta}} \left [ J (s,a,r; \theta, \Theta) \right ]$ for various deep reinforcement learning algorithms are delineated in table \ref{tab: obj_funcs}. In a value-based algorithm, such as the deep Q network (DQN), the neural network represents a function approximator for the Q function. 
Q function represents the expected return when the agent takes action $a_t$ in state $s_t$ and is defined as
$ Q(s, a) = \mathbb{E}_{\pi} \left [ \sum_{m} \gamma^m r_{m+t+1} | s_t = s, a_t = a \right]$ where $\gamma \in [0,1]$ is the discount factor and $\mathbb{E}_{\pi}[\cdots]$ denotes the expected value given that the agent follows the policy $\pi$. 
The policy refers to taking the action corresponding to the highest Q value. In policy based algorithms like advantage actor-critic (A2C), trust region policy optimization (TRPO) and proximal policy optimization (PPO). 
The policy is directly modeled as a neural network that maps the state $s$ to the corresponding optimal action $a$. This network is often integrated with a value network, which maps the state $s$ to its corresponding value $V(s)$. 
The value function is the expected future return for a particular state $s_t$ and is defined as $V(s) = \mathbb{E}_{\pi} \left [ \sum_{m} \gamma^m r_{m+t+1} | s_t = s \right]$. 
The policy network objective function corresponds to advantage weighted log-likelihood of chosen actions, where advantage function is defined as the difference between Q-function and value function. 
Algorithms such as TRPO and PPO employ importance sampling to correct for the estimation of the advantage function according to the old policy $\pi_{\theta_{old}}$ (that is, the policy before it is updated in a given policy iteration). 
As a result, the policy objective function contains the ratio term $\textbf{r}(\theta)= \pi_{\theta}(a|s)/\pi_{\theta_{old}}(a|s)$. 
Subsequently, the objective function for the integrated network is the sum of policy objective function added and value loss term multiplied by value coefficient $c_v$.
In the TRPO algorithm, the destructive steps of large gradients often encountered in policy gradient algorithms such as A2C are avoided by penalizing the KL-divergence between old and new policies with the factor $\beta$.
In the PPO algorithm, this is achieved by clipping the ratio $\textbf{r}(\theta)$ between $1-\epsilon$ and $1+\epsilon$ for a small value of $\epsilon \in [0,1]$. Furthermore, the exploration in policy search is maximized by maximizing the entropy of the learned policy $S[\pi_{\theta}](s)$ and is added in the objective function with the entropy coefficient $c_e$.
\begin{table}[ht]
    \caption{Objective function  $\mathbb{E}_{s,a,r \sim p_{\theta}} [ J (s,a,r; \theta, \Theta) ]$ for different deep RL algorithms }
    \centering
    \begin{tabular}{l l}
        \hline
        & \\
         Algorithm & Objective function, \\ 
         & $\mathbb{E}_{s,a,r \sim p_{\theta}} [ J (s,a,r; \theta, \Theta) ]$ \\
         & \\
        \hline
        & \\
        DQN & $\mathbb{E}_{s,a,r \sim p_{\theta}} \left [ \left ( r + \gamma \max_{a'}Q_{\theta_{old}}(s',a') - Q_{\theta}(s,a)   \right )^2 \right ] $ \\
        (value network) & \\
        & \\
        \hline
        & \\
        A2C & $\mathbb{E}_{s,a,r \sim p_{\theta}} \left [ \log \pi_{\theta}(a|s) A(s,a) - c_v \left (  r + \gamma \max_{s'}V_{\theta_{old}}(s') - V_{\theta}(s) \right )^2 \right ] $ \\
        (policy + value network)& \\
        & \\
        \hline
        & \\
        TRPO & $\mathbb{E}_{s,a,r \sim p_{\theta}}  [ \textbf{r}(\theta) A(s,a) - \beta \textup{KL}[\pi_{\theta_{old}}(\cdot|a) ,  \pi_{\theta}(\cdot|a)]$ \\
        (policy + value network)& $- c_v \left (  r + \gamma \max_{s'}V_{\theta_{old}}(s') - V_{\theta}(s) \right )^2  ] $ \\
        & \\
        \hline
        & \\
        PPO & $\mathbb{E}_{s,a,r \sim p_{\theta}}  [ \min \left (\textbf{r}(\theta) A(s,a), \textup{clip} (\textbf{r}(\theta), 1-\epsilon, 1+\epsilon)A(s,a) \right )$\\ 
        (policy + value network)& $- c_v \left (  r + \gamma \max_{s'}V_{\theta_{old}}(s') - V_{\theta}(s) \right )^2 + c_e S[\pi_\theta](s)  ] $ \\
        & \\
        \hline
    \end{tabular}
    \label{tab: obj_funcs}
\end{table}

\section{Principle behind computational savings of MLMC estimator} \label{app: funda_mlmc}
Suppose that we estimate the expectation of the quantity $P^L(\omega)$ where $\omega$ is a random variable that follows the probability distribution $\Omega$ (that is, $\omega \sim \Omega
$). 
The Monte Carlo estimate of this quantity is given by $\widehat{\mathbb{E}}_{\Omega}^{MC}(P^L(\omega)) = N^{-1} \sum_{i=1}^{N} P^L(\omega_i)$. 
If $C$ and $V$, respectively, correspond to the computational cost and variance of the term $P^L(\omega_i)$, the cost of the estimator $\widehat{\mathbb{E}}_{\Omega}^{MC}(P^L(\omega))$ is $CN$ while its overall variance is $VN^{-1}$.
That is, to achieve an overall variance of $\epsilon^2$, we need to choose $N=\epsilon^{-2}V$ (that is, $N\propto V$).
Now, if we suppose that we have an approximation of $P^L(\omega)$ defined as $P^l(\omega)$ such that $\mathbb{V}_{\Omega}[P^l(\omega)] >>> \mathbb{V}_{\Omega}[P^L(\omega) - P^l(\omega)]$, the two-level Monte Carlo estimator can be written as $\widehat{\mathbb{E}}_{\Omega}^{2LMC}(P^L(\omega)) = N_l^{-1} \sum_{i=1}^{N_l} P^l(\omega_i) + N_L^{-1} \sum_{i=1}^{N_L} P^L(\omega_i) - P^l(\omega_i)$. 
If $C_L$ and $V_L$ are the computational cost and variance of the term $P^L(\omega_i) - P^l(\omega_i)$ while $C_l$ and $V_l$ are the computational cost and variance of the term $P^l(\omega_i)$.
The total cost of this two-level Monte Carlo estimator can be computed as $N_lC_l + N_LC_L$ where $N_l\propto V_l$ and $N_L\propto V_L$. Since, by definition, $V_l>>>V_L$, we can also conclude that $N_l>>>N_L$. In other words, if $C_l<<<C_L$, computational cost of two-level Monte Carlo estimate $\widehat{\mathbb{E}}_{\Omega}^{2LMC}(P^L(\omega))$ is much smaller than the Monte Carlo estimate $\widehat{\mathbb{E}}_{\Omega}^{MC}(P^L(\omega))$. The same concept can be extended to multilevel Monte Carlo instead of two-level Monte Carlo estimate.

\section{Implementation of multilevel PPO in stable baselines 3} \label{app: mlppo_implentation}
The multilevel PPO algorithm is implemented using the Stable Baselines3 (SB3) \citep{stable-baselines3} library, which is a set of reliable implementations of reinforcement learning algorithms in PyTorch. The codes for the multilevel implementation can be found in the fork: \url{https://github.com/atishdixit16/stable-baselines3}. In the following text, the implementation of the classical PPO in SB3 is explained in detail. Then it is followed by additional implementations corresponding to the multilevel PPO algorithm. 
\subsection{Classical PPO implementation in stable baselines 3}
\begin{figure}[ht]
    \centering
    \begin{tikzpicture}[line cap=round,line join=round,>=triangle 45,x=1cm,y=1cm, scale=1.0]
\clip(-7,-3) rectangle (7,3);

\draw[draw=black, fill=gray!20] (-2,1) rectangle (2,2);

\draw[draw=black, fill=gray!20] (-3.5,-0.2) rectangle (3.5,-2.9);

\large
\draw (0,1.5) node[anchor=center] {$\pi_{\theta}(a|s)$};
\draw (2.1,2.1) node[anchor=north west] {$a_m$};
\draw (-5.1,-0.9) node[anchor=north west] {$r_{m+1}$};
\draw (-5.1,-2.2) node[anchor=north west] {$s_{m+1}$};
\draw (-2.9,1.2) node[anchor=north west] {$r_m$};
\draw (-2.9,2.3) node[anchor=north west] {$s_m$};
\draw (0,2) node[anchor=south] {$\mathrm{Agent}$};
\draw (0,-0.1) node[anchor=south] {$\mathrm{Environment}, E$};
\node[draw, fill=white] (P) at (1.5,-2.2) {$\mathcal{P}(s_m, a_m)$};
\node[draw, fill=white] (R) at (-1.5,-0.9) {$\mathcal{R}(s_m, a_m, s_{m+1})$};
\node[draw, fill=white] (s) at (-1.5,-2.2) {$s_{m+1}$};

\draw (-6.5,-2.2) -- (-5.2,-2.2);
\draw [dashed] (-5.2, -0.4) -- (-5.2, -2.8);
\draw [->] (-5.5,1.2) -- (-2,1.2);
\draw [->] (-6.5,1.8) -- (-2,1.8);
\draw (-5.5,-0.9)-- (-5.5,1.2);
\draw (-6.5,-2.2)-- (-6.5,1.8);
\draw (2,1.5) -- (6,1.5);
\draw (6,1.5)-- (6,-2.2);
\draw (-5.2,-0.9)-- (-5.5,-0.9);

\draw[->] (P.west) -- (s.east);
\draw[->] (s.north) -- (R.south);
\draw[->] (6,-2.2) -- (P.east);
\draw[->] (s.west) -- (-5.2,-2.2);
\draw[->] (R.west) -- (-5.2,-0.9);


\end{tikzpicture}
    \caption{A typical agent-environment interaction for classical framework}
    \label{fig: rl_framework_classical}
\end{figure}
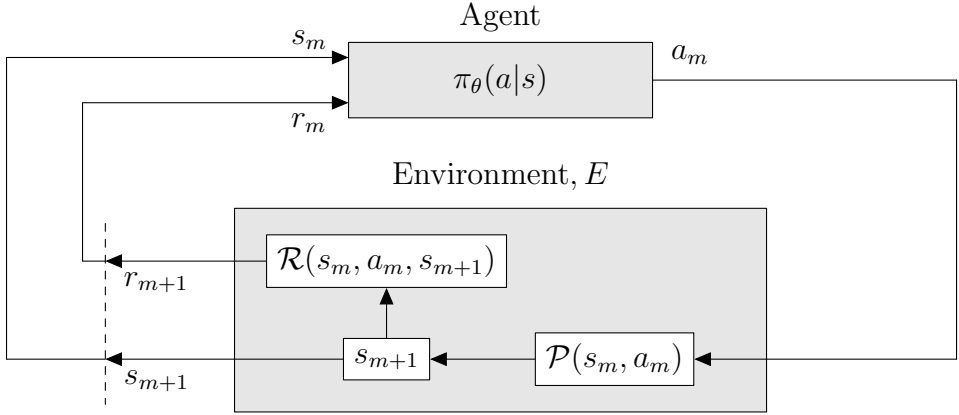

 RL framework consists of the environment $\mathcal{E}$ which is governed by a Markov decision process described by the tuple $\left \langle \mathcal{S},\mathcal{A},\mathcal{P},\mathcal{R}, \mu \right \rangle$.
Here, $\mathcal{S} \subset \mathbb{R}^{n_s}$ is the state-space, $\mathcal{A} \subset \mathbb{R}^{n_a}$ is the action-space, 
$\mathcal{P}(s'|s,a)$ is a Markov transition probability function between the current state $s$ and the next state $s'$ under action $a$ and $\mathcal{R}(s,a,s')$ is the reward function.
The function $\mu(s)$ returns a state from the initial state distribution if $s$ is the terminal state of the episode; otherwise, it returns the same state $s$. 
The goal of reinforcement learning is to find the policy $\pi_{\theta}(a|s)$ to take an optimal action $a$ when in the state $s$, by exploring the state-action space with what are called agent-environment interactions. Figure \ref{fig: rl_framework_classical} shows a typical schematic of such agent-environment interaction. The term \textit{agent} refers to the controller that follows the policy $\pi_{\theta}(a|s)$ while the \textit{environment} consists of the transition function, $\mathcal{P}$, and the reward function, $\mathcal{R}$.

The algorithm \ref{alg:ppo_stable_baselines} delimits the simplified implementation of the PPO algorithm in SB3. The algorithm's inputs are: environment $E$, number of actors $N$, number of steps in each policy iteration $T$, batch size $M$ ($\leq NT$) and number of epochs $K$. The data obtained through the rollouts of agent-environment interactions is stored in a buffer named RolloutBuffer in the format [$s, a, r, d, V, L_{\textup{old}}, R, A$], where the notation is
\begin{itemize}
    \item $s$: state,
    \item $a$: action,
    \item $r$: reward,
    \item $d$: episode terminal boolean (done),
    \item $V$: Value function (obtained from policy network rollout),
    \item $L_{\textup{old}}$: log probability value, $\log(\pi_{\theta_{old}}(a|s))$
    \item $R$: Return value (obtained using generalized advantage estimation),
    \item $A$: Advantage function (obtained using generalized advantage estimation).
\end{itemize}
 RolloutBuffer accumulates in total $N \times T$ rows of the above data in each iteration.
At the beginning of each iteration, the function \textbf{CollectRollouts} is used to fill in the data in RolloutBuffer. 
The total of $N \times T$ data rows is divided into batches of size $M$, each using the function \textbf{GetBatches}. 
The actor loss term $L_a$, the value loss term $L_v$ and the entropy loss term $L_e$ (defined in equation \ref{eq: ppo_obj}) are calculated for each such batch using the function \textbf{ComputeBatchLosses}. 
Finally, a Monte Carlo estimate for the loss term is computed as follows.
\begin{equation*}
    \textup{loss}_\textup{MC} = mean [L_a + L_v + L_e],
\end{equation*} 
which is used to update the policy parameters using automatic differentiation.
This is done using the function \textbf{UpdatePolicy} and is performed $K$ times for every batch.

\begin{algorithm}
	\caption{PPO implementation in stable baselines} 
	\begin{algorithmic}[1]
	\State Input: $E, N, T, M, K$
	\State $E.reset()$
	\State Generate empty RolloutBuffer
	\For{iteration, $i=1,2,\ldots$}
	    \State \textbf{CollectRollouts}($E$, $N$, $T$, RolloutBuffer)
	    \For {$epoch=1,2,\ldots,K$}
			\For {batch in \textbf{GetBatches}(RolloutBufferArray, $M$):}
			    \State $L_a$, $L_v$, $L_e$ = \textbf{ComputeBatchLosses}($\textup{batch}$)
			    \State $\textup{loss}_\textup{MC} = mean \left [ L_a + L_v + L_e  \right]$
				\State \textbf{UpdatePolicy}( $\textup{loss}_\textup{MC}$ )
			\EndFor
		\EndFor
	\EndFor
	\end{algorithmic} 
	\label{alg:ppo_stable_baselines}
\end{algorithm}

\begin{algorithm}
	\caption{\textbf{CollectRollouts}($E, N, T$, RolloutBuffer)} 
	\begin{algorithmic}[1]
	\State Information: a RolloutBuffer consists of following data: [$s, a, r, d, V, L_{\textup{old}}, R, A$]
	\State reset RolloutBuffer (i.e. empty the buffer)
	\For {t in range($T$):}
		    \State rollout current state $s$, through policy network to obtain $a$, $V$, $L_{\textup{old}}(a)$ on $N$ actors
		    \State if $s$ is terminal, $s=E.reset()$
		    \State $ s', r, d, \cdot = E.step(a)$ on $N$ actors 
		    \State compute $R$ and $A$ using GAE
		    \State add [$ s, a, r, d, V, L_{\textup{old}}, R, A $] in the $\textup{RolloutBuffer}$
    \EndFor
	\end{algorithmic} 
	\label{alg:CollectRollouts_classical}
\end{algorithm}

The algorithm \ref{alg:CollectRollouts_classical} delineates the steps of the function \textbf{CollectRollouts}. 
For every timestep, the data is obtained using policy rollout, environment transition (using $step$ function) and generalized advantage estimation (GAE) computation on all $N$ actors and stored in the RolloutBuffer. 
Finally, \textbf{ComputeBatchLosses} function is illustrated in the algorithm \ref{alg:ComputeBatchLosses_classical}. 
The algorithm lists steps to compute actor loss term $L_a$, value loss term $L_v$ and entropy loss term $L_e$ for the given batch. 
Note that the loss terms are the vectors of dimension $M$, which are added later, and its mean is treated as the final loss term. 
The mean function in this process indicates the \textit{Monte Carlo} estimator of the PPO loss term.

\begin{algorithm}
	\caption{\textbf{ComputeBatchLosses}(batch)} 
	\begin{algorithmic}[1]
	\State Information: a batch consists of $M$ rows following data: [$s, a, V, L_{\textup{old}}, R, A$]
	\State compute $V_{\textup{now}}$ and $L_{\textup{now}}(a)$ by rolling out $s$ through policy network
	\State compute ratio, $r_t = \exp{(L_{\textup{now}} - L_{\textup{old}})}$
	\State compute $L_1 = A r_t$ and $L_2 = A [\textup{clip}(r_t, 1-\epsilon, 1+\epsilon)]$
	\State $L_a = \min{(L_1, L_2)}$
	\State $L_v = C_v|V_{\textup{now}} - R |^2$ ($C_v$ is value loss term coefficient)
	\State $L_e = -C_eL_{\textup{now}} $ ($C_e$ is entropy loss term coefficient)
	\State \Return{$L_a$, $L_v$, $L_e$}
	\end{algorithmic} 
	\label{alg:ComputeBatchLosses_classical}
\end{algorithm}

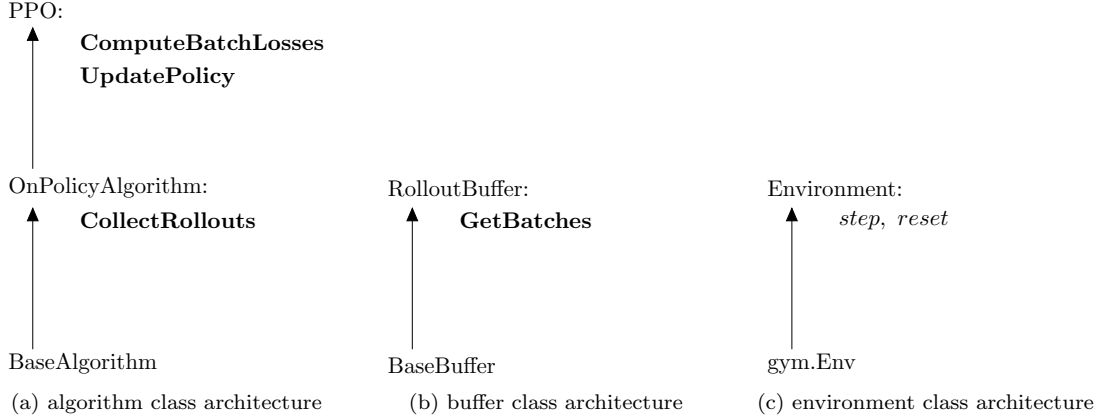
\begin{figure}
    \centering
    \begin{tabular}{c c c}
        \subfloat[algorithm class architecture ] {\resizebox{0.3\columnwidth}{!} {
        \begin{tikzpicture}[line cap=round,line join=round,>=triangle 45,x=1cm,y=1cm, scale=1.0]
\clip(-3,-3) rectangle (2,3);

\draw (-3,3) node[anchor=north west] {PPO: };
\draw (-1.9,2.5) node[anchor=north west] { \textbf{ComputeBatchLosses}};
\draw (-1.9,2) node[anchor=north west] { \textbf{UpdatePolicy}};

\draw (-3,0) node[anchor=west] {OnPolicyAlgorithm:};
\draw (-1.9,-0.5) node[anchor= west] { \textbf{CollectRollouts}};

\draw (-3,-3) node[anchor=south west] {BaseAlgorithm};

\draw [->,line width=0.4pt] (-2.5,-2.5) -- (-2.5,-0.3);
\draw [<-,line width=0.4pt] (-2.5,2.5) -- (-2.5,0.3);

\end{tikzpicture}
        }}  &
        \subfloat[buffer class architecture] {\resizebox{0.3\columnwidth}{!}{
        \begin{tikzpicture}[line cap=round,line join=round,>=triangle 45,x=1cm,y=1cm, scale=1.0]
\clip(-3,-3) rectangle (2,3);


\draw (-3,0) node[anchor=west] {RolloutBuffer:};
\draw (-1.9,-0.5) node[anchor= west] { \textbf{GetBatches}};

\draw (-3,-3) node[anchor=south west] {BaseBuffer};

\draw [->,line width=0.4pt] (-2.5,-2.5) -- (-2.5,-0.3);

\end{tikzpicture}
        }} &
        \subfloat[environment class architecture] {\resizebox{0.3\columnwidth}{!}{
        \begin{tikzpicture}[line cap=round,line join=round,>=triangle 45,x=1cm,y=1cm, scale=1.0]
\clip(-3,-3) rectangle (2,3);


\draw (-3,0) node[anchor=west] {Environment:};
\draw (-1.9,-0.5) node[anchor= west] {$step,\ reset$};

\draw (-3,-3) node[anchor=south west] {gym.Env};

\draw [->,line width=0.4pt] (-2.5,-2.5) -- (-2.5,-0.3);

\end{tikzpicture}
        }}
    \end{tabular}
    \caption{Object-oriented design for the stable baselines implementation of PPO algorithm }
    \label{fig:OOD}
\end{figure}

The class inheritance schema used in this implementation is shown in Figure \ref{fig:OOD}. The stable baselines use some more classes like Policy, Callbacks etc. but we present only the ones relevant to this discussion. \textbf{CollectRollouts} function belongs to OnPolicyAlgorithm which is the child of the BaseAlgorithm class and the parent of the PPO class. 
The functions \textbf{ComputeBatchLosses} and \textbf{UpdatePolicy} belong to the PPO class. 
BaseBuffer is the parent class for the RolloutBuffer class that contains the function \textbf{GetBatches}.
The Environment class (which is a child of the gym.Env class) contains functions such as $step$ and $reset$ corresponding to the transition function $\mathcal{P}$ and the initial state function $\mu$, respectively.

\subsection{Multilevel PPO implementation in stable baselines 3}
\begin{figure}[ht]
    \centering
    \begin{tikzpicture}[line cap=round,line join=round,>=triangle 45,x=1cm,y=1cm, scale=1.0]
\clip(-7,-7) rectangle (7,3);

\large

\draw[draw=black, fill=gray!20] (-2,1) rectangle (2,2);
\draw (0,2) node[anchor=south] {$\mathrm{Agent}$};
\draw (0,1.5) node[anchor=center] {$\pi_{\theta}(a^L|s^L)$};
\draw (-3.6,1.2) node[anchor=north west] {$r^l_m, \tilde{r}^{l-1}_m$};
\draw (-2.9,2.5) node[anchor=north west] {$s^L_m$};
\draw (2.1,2.3) node[anchor=north west] {$a^L_m$};

\draw[draw=black, fill=gray!20] (-3.5,-0.2) rectangle (3.5,-2.9);
\draw (0,-0.1) node[anchor=south] {$\mathrm{Environment}, E^l$};
\node[draw, fill=white] (P) at (1.5,-2.2) {$\mathcal{P}^l(s^l_m, a^l_m)$};
\node[draw, fill=white] (R) at (-1.5,-0.9) {$\mathcal{R}^l(s^l_m, a^l_m, s^l_{m+1})$};
\node[draw, fill=white] (s) at (-1.5,-2.2) {$s^l_{m+1}$};
\node[circle, fill=white, draw=black, inner sep=0, minimum size=0.8cm] (P_a) at (3.5,-2.2) {$\phi$};
\draw[->] (P.west) -- (s.east);
\draw[->] (s.north) -- (R.south);
\draw[->] (P_a.west) -- (P.east);
\draw[->] (6,-2.2) -- (P_a.east);
\draw[->] (s.west) -- (-5.2,-2.2);
\draw[->] (R.west) -- (-5.2,-0.9);
\node[circle, fill=white, draw=black, inner sep=0, minimum size=0.8cm] (P_s) at (-3.5,-2.2) {$\psi$};
\draw (-5.1,-0.9) node[anchor=north west] {$r^l_{m+1}$};
\draw (-5.1,-2.2) node[anchor=north west] {$s^L_{m+1}$};
\draw [dashed] (-5.2, -0.4) -- (-5.2, -2.8);

\draw[draw=black, fill=gray!20] (-3.5,-4.2) rectangle (3.5,-6.9);
\draw (0,-4.1) node[anchor=south] {$\mathrm{Synchronised\ Environment}, E^{l-1}.map\_from(E^l)$};
\node[draw, fill=white] (P) at (1.5,-6.2) {$\mathcal{P}^l(\tilde{s}^{l-1}_m, \tilde{a}^{l-1}_m)$};
\node[draw, fill=white] (R) at (-1.5,-4.9) {$\mathcal{R}^l(\tilde{s}^{l-1}_m, \tilde{a}^{l-1}_m, \tilde{s}^{l-1}_{m+1})$};
\node[draw, fill=white] (s) at (-1.5,-6.2) {$\tilde{s}^{l-1}_{m+1}$};
\node[circle, fill=white, draw=black, inner sep=0, minimum size=0.8cm] (P_a) at (3.5,-6.2) {$\phi$};
\draw[->] (P.west) -- (s.east);
\draw[->] (s.north) -- (R.south);
\draw[->] (P_a.west) -- (P.east);
\draw[->] (6,-6.2) -- (P_a.east);
\draw[->] (R.west) -- (-5.2,-4.9);
\draw (-5.1,-4.9) node[anchor=north west] {$\tilde{r}^{l-1}_{m+1}$};
\draw [dashed] (-5.2, -2.8) -- (-5.2, -5.8);
\draw (6,-2.2)-- (6,-6.2);

\draw (-6.5,-2.2) -- (-5.2,-2.2);
\draw [->] (-5.5,1.2) -- (-2,1.2);
\draw [->] (-6.5,1.8) -- (-2,1.8);
\draw (-5.5,-0.9)-- (-5.5,1.2);
\draw (-6.5,-2.2)-- (-6.5,1.8);
\draw (2,1.5) -- (6,1.5);
\draw (6,1.5)-- (6,-2.2);
\draw (-5.2,-0.9)-- (-5.5,-0.9);

\end{tikzpicture}
    \caption{A typical agent-environment interaction for an environment on level $l$ synchronized with environment on level $l-1$}
    \label{fig: rl_framework_ml}
\end{figure}
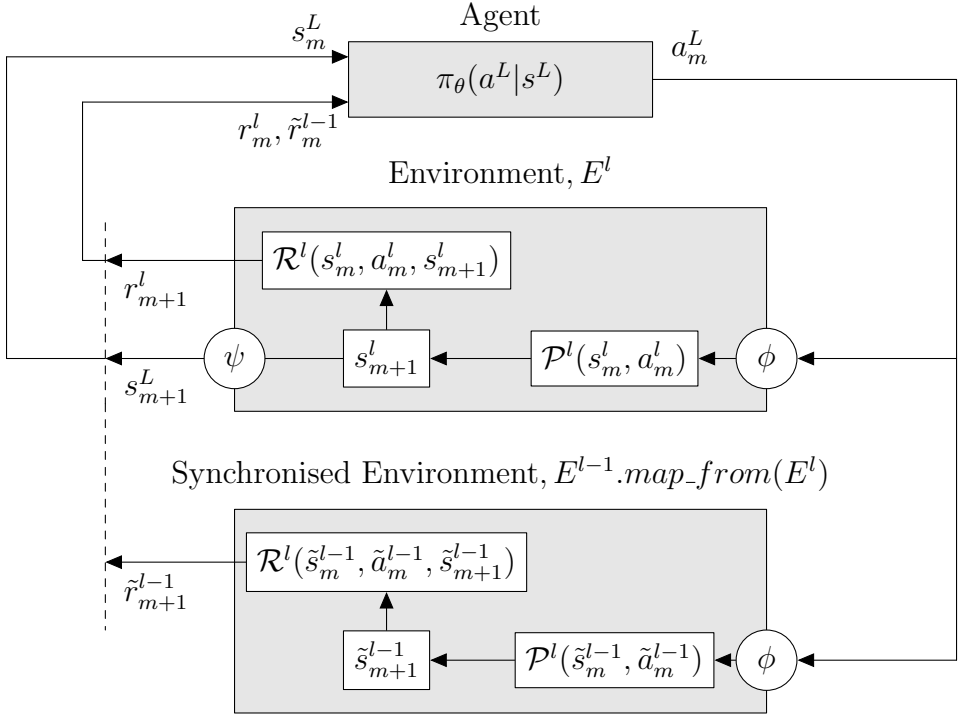

Figure \ref{fig: rl_framework_ml} illustrates a typical agent-environment interaction in multilevel PPO implementation. 
Multiple levels of environment are represented with $E^1, E^2, \ldots, E^{L-1}$ so that the computational cost of $\mathcal{P}^l$ and the accuracy of $\mathcal{R}^l$ are lower than $\mathcal{P}^{l+1}$ and $\mathcal{R}^{l+1}$, respectively. 
The environment corresponding to the grid fidelity factor $l$ consists of a transition function $\mathcal{P}_l$, which is achieved by discretizing the dynamical system, and a reward function $\mathcal{R}_l$. The policy network is designed with states $s^L$ and controls $a^L$, corresponding to the environment $E^L$. 
As a result, state $s^l_{m+1}$, in the environment, $E^l$ passes through the mapping $\psi^L_l$ which maps the state from level $l$ to level $L$.
Similarly, the action obtained from the policy network is passed through a mapping operator $\phi^l_L$, which maps the action from the level $L$ to the level $l$.

Algorithm \ref{alg:ppo_multilevel} illustrates the pseudocode for multilevel implementation of the PPO algorithm in the stable baselines library. The inputs are the same as in classical PPO implementation except multilevel variables are provided as an array of length $L$: environments at each level $\boldsymbol{E}=[E^1, E^2.\ldots E^L]$, number of actors $N$, number of steps in each level $\boldsymbol{T}=[T^1, T^2, \ldots, T^L]$, number of batches in each level $\boldsymbol{M}=[M^1, M^2, \ldots, M^L]$ (such that $NT^l \leq M^l$ and $T^1/M^1=\cdots=T^L/M^L$) and number of epochs $K$.
In multilevel implementation, we formulate the loss term's estimate using multilevel Monte Carlo which is given as
\begin{equation*}
    \textup{loss}_\textup{MLMC} =  \sum_{l=1}^L mean \left [ (L^l_a - \tilde{L}^{l-1}_a) + (L^l_v - \tilde{L}^{l-1}_v) + (L^l_e - \tilde{L}^{l-1}_e )  \right],
\end{equation*}
where $\tilde{L}^0_a, \tilde{L}^0_v$ and $\tilde{L}^0_e$ are set to zero. 
The outline of a typical agent-environment interaction to obtain synchronized samples of levels $l$ and $l-1$ is illustrated in Figure \ref{fig: rl_framework_ml}.
We use arrays of RolloutBuffers for each level, and each $\textup{RolloutBuffer}^l$ that collects rollouts at level $l$ has a synchronized buffer $\textup{SyncRolloutBuffer}^l$ that collects corresponding synchronized data at level $l-1$. This is achieved using the function \textbf{CollectRollouts}. 
Figure \ref{fig: rolloutBufferArray} illustrates the RolloutBufferArray and SyncRolloutBufferArray used in this algorithm. Furthermore, the \textbf{GetBatches} function is used to generate an array of batches, which is used to compute the multilevel Monte Carlo estimate of the loss term. 
The batch array consists of in total $NT^L/M^L$ batches, where each batch consists of $L$ batches from RolloutBuffers and $L$ batches from SyncRolloutBuffers.
Figure \ref{fig:batch_array} illustrates the batch array used in the algorithm. 
The $\textup{batch}^l$, $\textup{syncBatch}^{l-1}$ from $\textup{RolloutBuffer}^l$, $\textup{SyncRolloutBuffer}^l$ are used to compute the $\textup{loss}_\textup{MLMC}$ terms on the level $l$. In every batch, these terms are computed at each level and added to obtain $\textup{loss}_\textup{MLMC}$, which is used to update the policy network parameters using the function \textbf{UpdatePolicy}.

\begin{figure}
    \centering
    \begin{align*}
    \begin{bmatrix}
     \textup{RolloutBuffer}^1 \\
     \textup{RolloutBuffer}^2\\
     \cdot\\ 
     \cdot\\
     \textup{RolloutBuffer}^L\\
    \end{bmatrix}
    &
    \begin{bmatrix}
     \textup{SyncRolloutBuffer}^1\\
     \textup{SyncRolloutBuffer}^2\\
     \cdot\\ 
     \cdot\\
     \textup{SyncRolloutBuffer}^L\\
    \end{bmatrix}
    \end{align*}
    \caption{RolloutBufferArray (on left) and SyncRolloutBufferArray (on right).  $\textup{SyncRolloutBuffer}^l$ consists of synchronized data of $\textup{RolloutBuffer}^l$ with level $l$ to a level $l-1$. Each buffer with level $l$ consists of $N \times T_l$ rows of data in [$s, a, r, d, V, L_{\textup{old}}, R, A$] format. }
    \label{fig: rolloutBufferArray}
\end{figure}

\begin{figure}
    \centering
    \begin{align*}
    \begin{bmatrix}
        \begin{bmatrix}
            \textup{batch}^1, \textup{syncBatch}^0\\ 
            \textup{batch}^2, \textup{syncBatch}^1\\ 
            \cdot\\ 
            \cdot\\ 
            \textup{batch}^L, \textup{syncBatch}^{L-1}
        \end{bmatrix} & 
        \begin{bmatrix}
            \textup{batch}^1, \textup{syncBatch}^0\\ 
            \textup{batch}^2, \textup{syncBatch}^1\\ 
            \cdot\\ 
            \cdot\\ 
            \textup{batch}^L, \textup{syncBatch}^{L-1}
        \end{bmatrix} & 
        \begin{bmatrix}
            \cdots\\ 
            \cdots\\ 
            \cdots\\ 
            \cdots\\ 
            \cdots
        \end{bmatrix} &
        \begin{bmatrix}
            \textup{batch}^1, \textup{syncBatch}^0\\ 
            \textup{batch}^2, \textup{syncBatch}^1\\ 
            \cdot\\ 
            \cdot\\ 
            \textup{batch}^L, \textup{syncBatch}^{L-1}
        \end{bmatrix} 
    \end{bmatrix}
    \end{align*}
    \caption{$\textup{batch\_array}$ which is achieved from \textbf{GetBatches} function. It consists of in total $NT_l/M_l$ batches as shown with the columns of the array. Each such batch consists of $L$ batches from RolloutBuffers (denoted by $\textup{batch}^{l}$) and SyncRolloutBuffers (denoted by $\textup{syncBatch}^{l-1}$). $\textup{batch}^l$ and $\textup{SyncBatch}^{l-1}$ consists of $M^l$ rows of data in the format, [$o, a, V, L_{\textup{old}}, R, A$]. }
    \label{fig:batch_array}
\end{figure}

\begin{algorithm}
	\caption{Multilevel proximal policy optimization pseudocode} 
	\begin{algorithmic}[1]
	\State Input: $\boldsymbol{E}, N, \boldsymbol{T}, \boldsymbol{M}, K$
	\State $E^1.reset()$
	\State Generate empty RolloutBufferArray, SyncRolloutBufferArray
	\For{iteration, $i=1,2,\ldots$}
	    \State \textbf{CollectRollouts}($\boldsymbol{E}, N, \boldsymbol{T}$, RolloutBufferArray, SyncRolloutBufferArray)
	    \For {$epoch=1,2,\ldots,K$}
			\For {batch\_array in \textbf{GetBatches}(RolloutBufferArray, SyncRolloutBufferArray, $\boldsymbol{M}$):}
				\State $\textup{loss}_\textup{MLMC}=0$
				\For{ $\textup{batch}^l, \textup{syncBatch}^{l-1}$ in batch\_array}
				    \State $L^l_a$, $L^l_v$, $L^l_e$ = \textbf{ComputeBatchLosses}($\textup{batch}^l$)
				    \If{$l > 1$}
                        \State $\tilde{L}^{l-1}_a$, $\tilde{L}^{l-1}_v$, $\tilde{L}^{l-1}_e$ = \textbf{ComputeBatchLosses}($\textup{syncBatch}^{l-1}$)
                    \Else
                        \State $\tilde{L}^{l-1}_a$, $\tilde{L}^{l-1}_v$, $\tilde{L}^{l-1}_e$ = 0
                    \EndIf
				    \State $L^l = mean \left [ (L^l_a - \tilde{L}^{l-1}_a) + (L^l_v - \tilde{L}^{l-1}_v) + (L^l_e - \tilde{L}^{l-1}_e ) \right]$
				    \State $\textup{loss}_\textup{MLMC}$ = $\textup{loss}_\textup{MLMC}$ + $L^l$
				\EndFor
				\State \textbf{UpdatePolicy}( $\textup{loss}_\textup{MLMC}$)
			\EndFor
		\EndFor
	\EndFor
	\end{algorithmic} 
	\label{alg:ppo_multilevel}
\end{algorithm}

The algorithm \ref{alg:CollectRollouts} delimits the function \textbf{CollectRollouts} used in multilevel implementation.
At each level $l$ the $\textup{RolloutBuffer}^l$ is filled with the data, and the corresponding synchronized data at the level $l-1$ is filled in the $\textup{SyncRolloutBuffer}^l$.
Since $L^0_a$, $L^0_v$ and $L^0_e$ are set to zero, the data in $\textup{SyncRolloutBuffer}^1$ are filled with None values.
The mapping functions $\psi_l^{l'}$ and $\phi_l^{l'}$ are implemented as a set of functions in the definition of the environment $E^l$.
As a result, the mapping of state ($\psi_l^L$ from Equation \ref{eq: ml_rollout_notation}) and action ($\phi_L^l$ from equation \ref{eq: ml_rollout_notation}) to and from the policy + value network is denoted with shorthand notation $\psi$ and $\phi$, respectively.
Synchronization of state from level $l$ to $l'$ is indicated by $map\_from$ function that maps an environment $E^l$ to another environment at level $l'$, denoted as $E^{l'}$.
Algorithm \ref{alg:GetBatches} illustrates the pseudocode for the \textbf{GetBatches} function, which creates mini-batches (as illustrated in Figure \ref{fig:batch_array}) from collected data in RolloutBufferArray and SyncRolloutBufferArray.

\begin{algorithm}
	\caption{\textbf{CollectRollouts}($\boldsymbol{E}, N, \boldsymbol{T}$, RolloutBufferArray, SyncRolloutBufferArray)} 
	\begin{algorithmic}[1]
	\State Information: a RolloutBuffer consists of following data: [$s, a, r, d, V, L_{\textup{old}}, R, A$]
	\State reset RolloutBufferArray, SyncRolloutBufferArray (i.e. empty the buffers)
	\For{ $T^l, E^l, \textup{RolloutBuffer}^l, \textup{SyncRolloutBuffer}^l$ in $\boldsymbol{E}, \boldsymbol{T}$, RolloutBufferArray, SyncRolloutBufferArray}
	    \If{$l > 1$}
            \State $E^{l}.map\_from(E^{l-1})$
        \EndIf
		\For {t in range($T^l$):}
            \State $s^l=E^l.reset()$ if $s^l$ is terminal
            \State $s^L=E^l.\psi(s^l)$
            \State $a^L=\pi_{\theta}(a^L|s^L)$
            \State $a^l=\phi(a^L)$
		    \State compute $V^l$ and $L_{\textup{old}}(a^L)$
		    \State $ \cdot, r^l, d^l, \cdot= E^l.step(a^l)$ on $N$ actors
		    \State compute $R^l$ and $A^l$ using GAE
		    \State add [$ s^l, a^l, r^l, d^l, V^l, L_{\textup{old}}^l, R^l, A^l $] in the $\textup{RolloutBuffer}^l$
		    \State
		    \If{$l > 1$}
		        \State $ E^{l-1}.map\_from(E^{l})$
		        \State $\tilde{s}^L=E^{l-1}.\psi(\tilde{s}^{l-1})$
		        \State $\tilde{a}^{l-1} = a^l $
		        \State $\tilde{a}^L=\pi_{\theta}(\tilde{a}^L|\tilde{s}^L)$
		        \State $\tilde{a}^{l-1}=E^{l-1}.\phi(\tilde{a}^L)$
		        \State compute $\tilde{V}^{l-1}$ and $\tilde{L}_{\textup{old}}(a^{L})$
		        \State $ \cdot, \tilde{r}^{l-1}, \cdot, \cdot= E^{l-1}.step(\tilde{a}^{l-1})$ on $N$ actors
		        \State compute $\tilde{R}^{l-1}$ and $\tilde{A}^{l-1}$ using GAE
		        \State add [$ \tilde{s}^{l-1}, \tilde{a}^{l-1}, \tilde{r}^{l-1}, \tilde{d}^{l}, \tilde{V}^{l-1}, \tilde{L}_{\textup{old}}^{l-1}, \tilde{R}^{l-1}, \tilde{A}^{l-1} $] in the $\textup{SyncRolloutBuffer}^l$
            \Else 
                \State add [ None, $\ldots$, None] in the $\textup{SyncRolloutBuffer}^l$
            \EndIf
		\EndFor
	\EndFor
	\end{algorithmic} 
	\label{alg:CollectRollouts}
\end{algorithm}

\begin{algorithm}
	\caption{\textbf{GetBatches}(RolloutBufferArray, SyncRolloutBufferArray, $\boldsymbol{M}$)} 
	\begin{algorithmic}[1]
	\State set $\textup{batch\_array}$ to an empty array 
	\For{$\textup{RolloutBuffer}^l, \textup{SyncRolloutBuffer}^l, M^l$ in RolloutBufferArray, SyncRolloutBufferArray, $\boldsymbol{M}$}
	   \State set batches to an empty array
	    \For{$\textup{batch}^{l}$, $\textup{batch}^{l-1}$ in \textbf{GetSyncBatches}($\textup{RolloutBuffer}^l$, $\textup{SyncRolloutBuffer}^l$, $M^l$) }
	        \State batches.append([$\textup{batch}^{l}$, $\textup{batch}^{l-1}$])
	    \EndFor
	   \State $\textup{batch\_array}.append(\textup{batches})$
	\EndFor
	\State \Return{$\textup{batch\_array}$}
	\end{algorithmic} 
	\label{alg:GetBatches}
\end{algorithm}

\begin{figure}
    \centering
    \begin{tabular}{c c c}
        \subfloat[algorithm class architecture ] {\resizebox{0.3\columnwidth}{!} {
        \begin{tikzpicture}[line cap=round,line join=round,>=triangle 45,x=1cm,y=1cm, scale=1.0]
\clip(-3,-3) rectangle (2,3);

\draw (-3,3) node[anchor=north west, text=red] {$\textup{PPO\_ML}$: };
\draw (-1.9,2.5) node[anchor=north west] { \textbf{ComputeBatchLosses}};
\draw (-1.9,2) node[anchor=north west] { \textbf{UpdatePolicy}};

\draw (-3,0) node[anchor=west, text=red] {OnPolicyAlgorithmMultilevel:};
\draw (-1.9,-0.5) node[anchor= west, text=red] { \textbf{CollectRollouts}};
\draw (-1.9,-1) node[anchor= west, text=red] { \textbf{GetBatches}};

\draw (-3,-3) node[anchor=south west] {BaseAlgorithm};

\draw [->,line width=0.4pt] (-2.5,-2.5) -- (-2.5,-0.3);
\draw [<-,line width=0.4pt] (-2.5,2.5) -- (-2.5,0.3);

\end{tikzpicture}
        }}  &
        \subfloat[buffer class architecture] {\resizebox{0.3\columnwidth}{!}{
        \begin{tikzpicture}[line cap=round,line join=round,>=triangle 45,x=1cm,y=1cm, scale=1.0]
\clip(-3,-3) rectangle (2,3);


\draw (-3,0) node[anchor=west] {RolloutBuffer:};
\draw (-1.9,-0.5) node[anchor= west, text=red] { \textbf{GetSyncBatches}};

\draw (-3,-3) node[anchor=south west] {BaseBuffer};

\draw [->,line width=0.4pt] (-2.5,-2.5) -- (-2.5,-0.3);

\end{tikzpicture}
        }} &
        \subfloat[environment class architecture] {\resizebox{0.3\columnwidth}{!}{
        \begin{tikzpicture}[line cap=round,line join=round,>=triangle 45,x=1cm,y=1cm, scale=1.0]
\clip(-3,-3) rectangle (2,3);


\draw (-3,0) node[anchor=west, text=red] {EnvironmentMultilevel :};
\draw (-1.9,-0.5) node[anchor= west, text=red] {$step,\ reset$};
\draw (-1.9,-1.0) node[anchor= west, text=red] {$ \psi, \phi, map\_from$};

\draw (-3,-3) node[anchor=south west] {gym.Env};

\draw [->,line width=0.4pt] (-2.5,-2.5) -- (-2.5,-0.3);

\end{tikzpicture}
        }}
    \end{tabular}
    \caption{Object-oriented design for the stable baselines implementation of multilevel PPO algorithm. The updated (from classical PPO implementation) definitions of functions and classes are highlighted in red colour. }
    \label{fig:OOD_ml}
\end{figure}
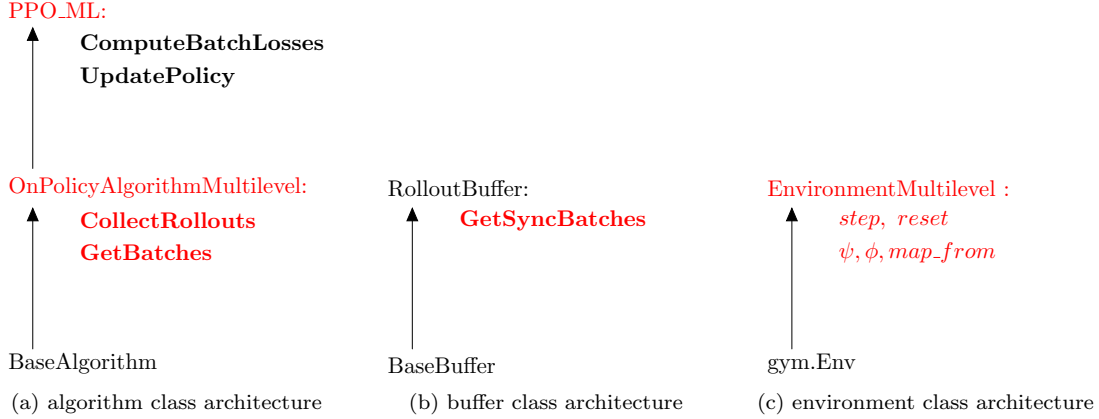

The class inheritance schema used in the multilevel implementation is shown in figure \ref{fig:OOD_ml}. \textbf{CollectRollouts} function belongs to OnPolicyAlgorithmMultilevel which is the child of BaseAlgorithm class and the parent of the $\textup{PPO\_ML}$ class. The functions \textbf{ComputeBatchLosses} and \textbf{UpdatePolicy} belong to the class $\textup{PPO\_ML}$. BaseBuffer is the parent class for the RolloutBuffer class that contains the function \textbf{GetBatches}.
The environment class architecture for multilevel framework is similar to that for classical framework except for the additional mapping functions $\psi$, $\phi$ and $map\_from$.
The updated definitions of the classes and functions are highlighted in red in figure \ref{fig:OOD_ml}.

\section{Cluster analysis of permeability uncertainty distribution} \label{app: cluster}
A set of permeability samples $\textbf{k}=\{k_1,\ldots,k_l\}$, is chosen to represent the variability in the permeability distribution $\mathcal{K}$. 
For the optimal control problem, our main interest is the uncertainty in the dynamical response of permeability, rather than the uncertainty in permeability itself. 
As a result, the connectivity distance \citep{park2011modeling} is used as a measure of the distance between the permeability field samples. 
The connectivity distance matrix $\textbf{D} \in \mathbb{R}^{N \times N}$ among the $N$ samples of $\mathcal{K}$ is formulated as
\begin{equation*}
    \textbf{D}(k_i, k_j) = \sum_{x''} \int_{t_0}^{T} \left [ c(x'', t; k_i) - c(x'', t; k_j) \right ]^2 dt,
    \label{eq: conn_dist}
\end{equation*}
where $N$ corresponds to a large number of samples of uncertainty distribution, $c(x'',t;k_i)$ is the concentration at the location $x''$ and at time $t$, when the permeability is set to $k_i$ and all wells are open equally. 
The multidimensional scaling of the distance matrix \textbf{D} is used to produce $N$ two-dimensional coordinates $d_1, d_2, \cdots, d_N$, each representing a permeability sample. 
The coordinates $d_1, d_2, \cdots, d_N$ are obtained such that the distance between $d_i$ and $d_j$ is equivalent to $\textbf{D}(k_i, k_j)$. In the k-means clustering process, these coordinates are divided into $l$ sets $S_1, S_2, \cdots , S_l$, obtained by solving the optimization problem:
\begin{equation*}
    \arg \min_{S} \sum_{i}^{l} \sum_{d_j \in S_i} \left \|   d_j - \mu_{S_i} \right \|,
\end{equation*}
where $\mu_{S_i}$ is the average of all coordinates in the set $S_i$. The training vector \textbf{k} is a set of $l$ samples of $\mathcal{K}$ where each of its values $k_i$ corresponds to the closest one to $\mu_{S_i}$.
\begin{figure*}
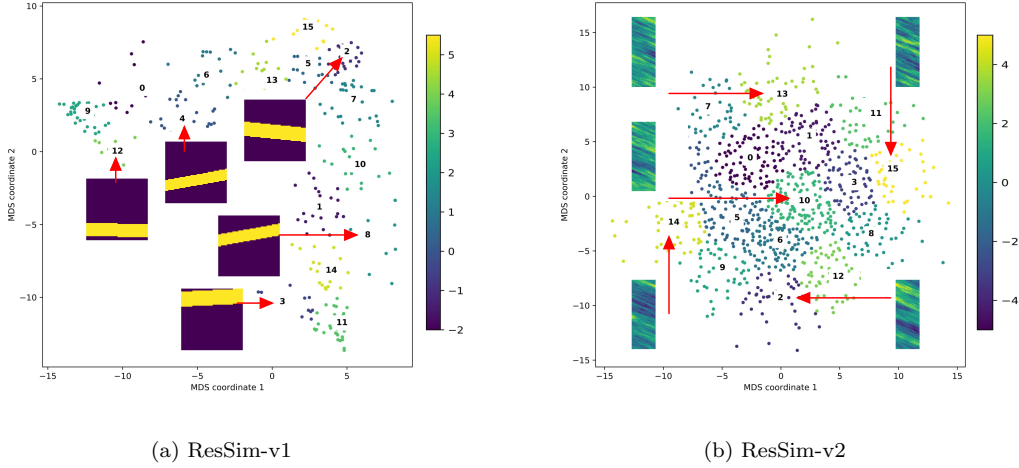

    \centering
    \begin{tabular}{cc}
    
        \subfloat[ResSim-v1] {\resizebox{0.45\textwidth}{!} {
        \input{images/case_1_cluster_image.tikz}
        }}  &
        \subfloat[ResSim-v2] {\resizebox{0.45\textwidth}{!}{
        \input{images/case_2_cluster_image.tikz}
        }}
    \end{tabular}
    \caption{clustering visualization for permeability samples }
    \label{fig: cluster}
\end{figure*}
The total number of samples $N$ and clusters $l$ is chosen to be 1000 and 16 for both uncertainty distributions, $\mathcal{G}_1$ and $\mathcal{G}_2$. 
A training vector $\textbf{k}$ is obtained with samples $k_1, \cdots, k_{16}$ each corresponding to a cluster center.
Figures \ref{fig: cluster}a and \ref{fig: cluster}b show cluster plots of permeability samples for ResSim-v1 and ResSim-v2. 

\section{Algorithm Parameters} \label{app: rl_params}
Parameters used for PPO are tabulated in Table \ref{tab: ppo_param} which were tuned using trial and error. 
For PPO algorithms, the parameters were essentially tuned to find the least variability in the learning plots.
The parameters of the DE algorithm are delineated in Table \ref{tab: de_param}. 
The code repository for both test cases presented in this article can be found at the link: \url{https://github.com/atishdixit16/multilevel_ppo}.

\begin{table}[ht]
    \caption{PPO algorithm parameters}
    \centering
    \begin{tabular}{l l l l}
        \hline
         & ResSim-v1 & ResSim-v2\\
        \hline
        discount rate, $\gamma$ & 0.99 & 0.99 \\
        clip range, $\epsilon$ & 0.1 & 0.15 \\
        policy network MLP layers & [93,150,100,80,62] & [35,70,70,50,21] \\
        policy network activation functions  & tanh & tanh\\
        policy network optimizers  & Adam & Adam \\
        learning rate & 3e-6 & 1e-5\\
        \hline
    \end{tabular}
    \label{tab: ppo_param}
\end{table}

\begin{table}[ht]
    \caption{DE algorithm parameters}
    \centering
    \begin{tabular}{l l l l}
        \hline
         & ResSim-v1 & ResSim-v2\\
        \hline
        number of CPUs  & 64 & 64 \\
        number of iterations & 1024 & 1024 \\
        population size & 310 & 105 \\
        recombination factor & 0.9 & 0.9 \\
        mutation factor & (0.5,1) & (0.5,1) \\
        \hline
    \end{tabular}
    \label{tab: de_param}
\end{table}

\clearpage
\bibliographystyle{plainnat}
\bibliography{references}

\end{document}